\documentclass[twoside,journey]{IEEEtran}
\usepackage{makecell}
\usepackage{hyperref}
\usepackage{array}
\usepackage{graphicx,amssymb,amsmath}
\usepackage{multicol}
\usepackage[noadjust]{cite}
\usepackage{setspace}
\usepackage{subcaption}
\usepackage{graphicx}
\usepackage{float}
\usepackage {url}
\usepackage{stfloats}
\usepackage{amsthm,pifont}
\usepackage{flushend}
\usepackage{cases,subeqnarray}
\usepackage{bm,multirow,bigstrut}
\usepackage{amsmath, amsthm, amssymb}
\usepackage{textcomp}
\usepackage{latexsym,bm}
\usepackage{booktabs}
\usepackage{xcolor}
\usepackage{mathtools}
\usepackage{dsfont}
\usepackage{extarrows}
\usepackage{epsfig}
\usepackage{epsfig}
\usepackage{epstopdf}
\usepackage[noend]{algpseudocode}
\usepackage{algorithmicx,algorithm}
\usepackage{makecell}
\usepackage{colortbl}
\usepackage{booktabs}
\usepackage{graphicx} % For scaling table to text width
\usepackage{array}    % For centering cells in a column
\usepackage{hyperref} % For hyperlinks
\usepackage{booktabs}
\usepackage{enumitem}
\usepackage{diagbox}
\usepackage{tikz}
\usetikzlibrary{trees,shadows.blur}
\usepackage{forest}
\usepackage{tabularx}
\usepackage{booktabs}
\usepackage{enumitem}
\theoremstyle{plain}

\theoremstyle{plain}

\usepackage{amsmath}

\newcommand{\ignore}[1]{{{\color{yellow} }}}
\renewcommand{\arraystretch}{1.5} % 调整行高
\definecolor{blue-green}{rgb}{0.0, 0.87, 0.87}
\IEEEoverridecommandlockouts
\begin{document}

%----------------------------title&author&thanks----------------------------
% \title{When Networks Meets Generative AI: Potentials, Challenges, and Directions}
\title{From Digital Twins to World Models: Opportunities, Challenges, and Applications for Mobile Edge General Intelligence}
\author{Jie~Zheng, Dusit Niyato, \emph{Fellow, IEEE}, Changyuan~Zhao, Jiawen~Kang, Jiacheng~Wang \\ %{\emph{(Invited paper)}} 

%\thanks{This work was supported in part by the National Natural Science Foundation of China (Grants nos. 61701400, 61501372, 61373176, and 61672426), by the Postdoctoral Foundation of China (Grants nos. 2017M613188 and 2017M613186)£¬ by Natural Science Special Foundation of Education Department in Shannxi (17JK0783).}
\thanks{J. Zheng is with State-Province Joint Engineering and Research Center of Advanced Networking and Intelligent Information Services, College of Computer, Northwest University,  Xian, 710127, Shaanxi, China.(jzheng@nwu.edu.cn) } %(Corresponding author: Jie Zheng.)}
\thanks{D. Niyato, C. Zhao, R. Zhang and J. Wang are with the College of Computing and Data Science, Nanyang Technological University, Singapore 639798. (dniyato@ntu.edu.sg, zhao0441@e.ntu.edu.sg, jiacheng.wang@ntu.edu.sg}
\thanks{J. Kang with the Automation of School, Guangdong University of Technology, Guangzhou 510006, China.(kavinkang@gdut.edu.cn)}
% \thanks{Z. Xiong is with the Queen's University Belfast, BT7 1NN, United Kingdom (z.xiong@qub.ac.uk)}
% \thanks{H. Zhang is with the Institute of Artificial Intelligence, University of Science and Technology Beijing, Beijing 100083, China. (zhanghaijun@ustb.edu.cn)}
% \thanks{A. Jamalipour is with the School of Electrical and Computer Engineering, University of Sydney, Australia (e-mail: a.jamalipour@ieee.org)}
% \thanks{D. I. Kim is with the Department of Electrical and Computer
% Engineering, Sungkyunkwan University, Suwon 16419, South Korea
% (email:dongin@skku.edu)}

}

\maketitle
%----------------------------abstract----------------------------
\vspace{-1cm}

\begin{abstract}
The rapid evolution toward 6G and beyond communication systems is accelerating the convergence of digital twins and world models at the network edge. Traditional digital twins provide high-fidelity representations of physical systems and support monitoring, analysis, and offline optimization. However, in highly dynamic edge environments, they face limitations in autonomy, adaptability, and scalability. This paper presents a systematic survey of the transition from digital twins to world models and discusses its role in enabling edge general intelligence (EGI). First, the paper clarifies the conceptual differences between digital twins and world models and highlights the shift from physics-based, centralized, and system-centric replicas to data-driven, decentralized, and agent-centric internal models. This discussion helps readers gain a clear understanding of how this transition enables more adaptive, autonomous, and resource-efficient intelligence at the network edge. The paper reviews the design principles, architectures, and key components of world models, including perception, latent state representation, dynamics learning, imagination-based planning, and memory. In addition, it examines the integration of world models and digital twins in wireless EGI systems and surveys emerging applications in integrated sensing and communications, semantic communication, air–ground networks, and low-altitude wireless networks. Finally, this survey provides a systematic roadmap and practical insights for designing world-model-driven edge intelligence systems in wireless and edge computing environments. It also outlines key research challenges and future directions toward scalable, reliable, and interoperable world models for edge-native agentic AI.
\end{abstract}
%----------------------------keywords----------------------------
%\newpage
\IEEEpeerreviewmaketitle
%----------------------------introduction----------------------------
%\input{outline}
\section{Introduction}\label{intro}

\subsection{Background}
Edge computing is undergoing a paradigm shift from task-specific edge artificial intelligence (Edge AI) to edge general intelligence (EGI)~\cite{EGI2025chen}. EGI represents a new class of intelligent systems deployed close to the physical world, capable of long-term autonomous operation on multiple tasks, environments, and time scales. EGI enables low-latency or even ultra-low-latency inference while reducing system energy consumption and operational costs, making it well suited for resource-constrained devices and diverse vertical applications~\cite{Syed2025edgeAisuv}. Traditional Edge AI is generally considered to have emerged from the development needs of interconnected ecosystems. Its primary objective is to enable the execution of local algorithms near data sources or edge servers, thus supporting applications with stringent requirements of low latency and high data efficiency, such as in-vehicle communication for autonomous driving~\cite{Katare2023edgeAIsuv}. Unlike traditional Edge AI, which focuses on predefined inference tasks, EGI agents can autonomously perceive, reason, and act within dynamic and partially observable environments.

By executing algorithms directly on edge devices, edge intelligence supports localized data processing and reduces reliance on cloud infrastructure. This approach improves latency and enhances privacy and security, as sensitive data remain within the local environment~\cite{10.1145/3724420}. Key application scenarios include UAV-enabled communication and sensing networks~\cite{Pandey2025uavSuv}, intelligent transportation systems with autonomous vehicles~\cite{Liu2026vehiclesSuv}, and industrial infrastructures for smart factories and energy grids~\cite{Balkus20225gVehiculaSuv}. In these scenarios, EGI is typically required to operate in close coupling with physical systems under resource constraints and in real time to address dynamic environments. This need for a closed loop of perception, decision-making, and action gives rise to clear embodied characteristics, aligning edge intelligence with the core principles of embodied artificial intelligence. Embodied intelligence emphasizes autonomous behavior achieved through perception and interaction with the physical world, and relies on physical simulators and world models to support training, environment representation, and predictive planning, thus advancing agents toward higher levels of autonomy~\cite{long2025survey}.

Achieving high levels of autonomy requires capabilities beyond simple reactive inference. Policies based only on instantaneous perception-to-action mappings typically implemented through end-to-end models are often fragile when faced with environmental changes, partial observability, and action delays. Thus, EGI systems must incorporate world models, which are internal representations of the external environment~\cite{ding2025understanding}. World models are commonly regarded as essential tools for understanding the current state of an environment and predicting its future evolution. By modeling state transitions, reasoning under uncertainty, and simulating and comparing different action sequences, world models support long-term planning and decision evaluation. World models can simulate complex real-world dynamics~\cite{goff2025learning}, while edge intelligence without such models often relies on short-term reactive decisions, making it challenging to ensure reliability and generalization in dynamic or unknown environments.

\begin{figure}[t!]
  \centering
  \includegraphics[width=\linewidth]{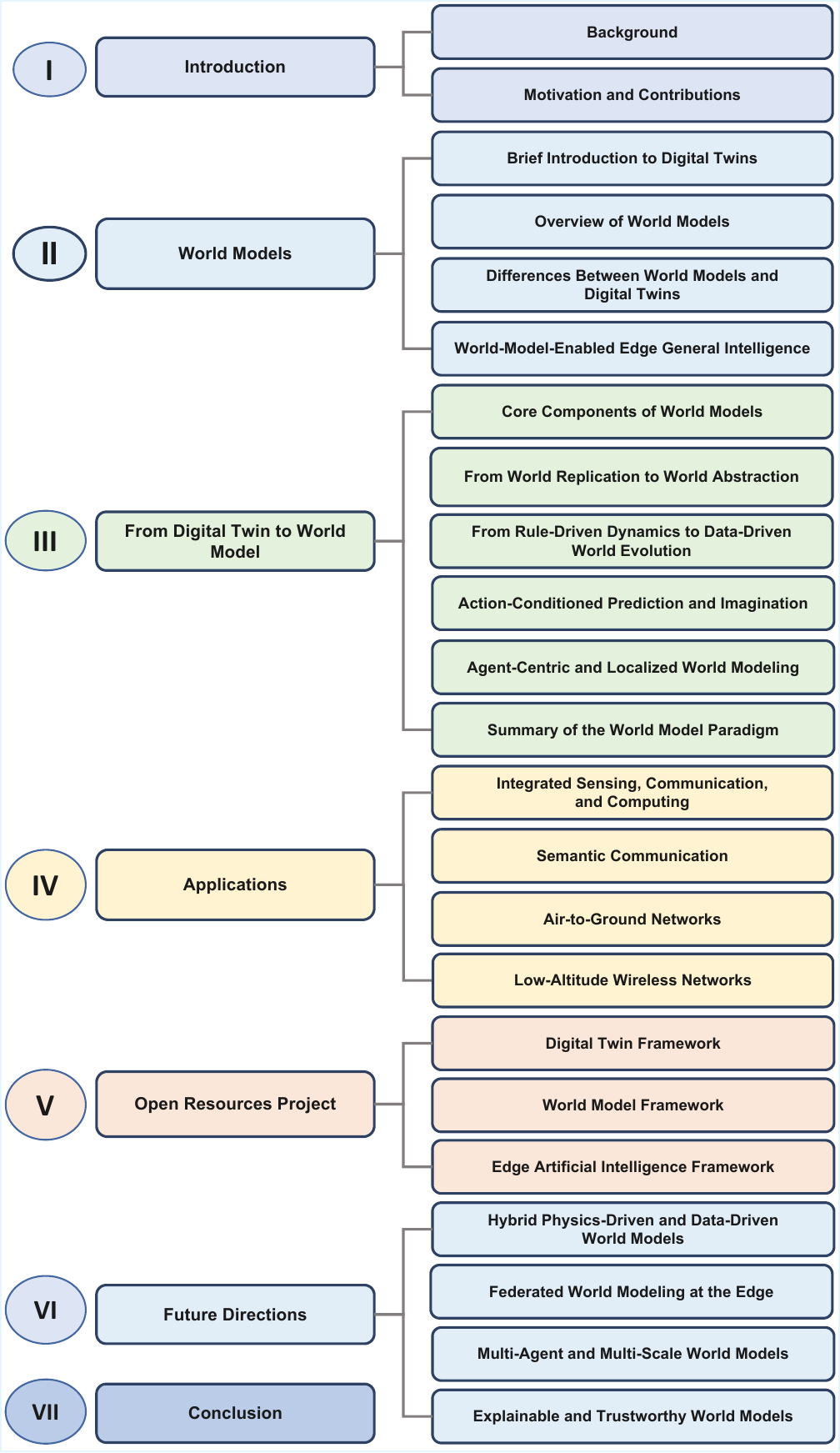}
  \caption{Conceptual architecture from digital twins to world models for edge general intelligence.}
  \label{fig:architecture}
  \vspace{-4mm}
\end{figure}

\subsection{Motivation and Contribution}
From early conceptual proposals to recent systematic developments, digital twins have become a main paradigm for representing and modeling physical objects in networked cyber–physical systems~\cite{Empl2025digitaltwinsSuv}. Digital twins are typically implemented in software as high-fidelity virtual counterparts of physical systems, supporting contextual modeling, real-time state synchronization, and secure modular interactions through application programming interfaces (APIs). Thus, digital twins can facilitate industrial analysis and decision-making tasks~\cite{10.1145/3772366}. Built on explicit physical laws, domain-specific equations, and expert rules, digital twins provide substantial value for offline engineering analysis and system design.

In recent years, digital twins have been widely applied in fields such as robotics, healthcare monitoring, and wireless communications. For example, RoboTwin integrates three-dimensional generative foundation models with large language models (LLMs) to generate diverse expert data for dual-arm robotic manipulation, leading to significant performance gains~\cite{Mu_2025_CVPR}. digital-twin-based models for smart home healthcare monitoring support visual monitoring, health state prediction, and intelligent control~\cite{10234399}. In wireless communication and edge computing systems, digital twins are used for network planning, configuration validation, and what-if analysis. They also enable joint optimization of communication and computing under low-latency and energy-constrained conditions while meeting the model accuracy requirements~\cite{10375758}.

However, digital twin technology still faces several challenges, including complex communication and data integration, limited data availability for machine learning training, high computational costs for high-fidelity modeling, strong dependence on interdisciplinary collaboration, and the lack of unified development and validation frameworks~\cite{Mihai2022DigitalTA}. Traditional digital twin paradigms, which are mainly designed for offline system engineering, are fundamentally misaligned with the requirements of online and autonomous EGI. This mismatch is further amplified in heterogeneous operational technology environments, where infrastructure differences and missing standards hinder large-scale deployment~\cite{Empl2025digitaltwinsSuv}. Moreover, digital twins often rely heavily on predefined rules and prior assumptions, which limits their generalizability. High-fidelity modeling also struggles to meet the real-time constraints of resource-limited edge devices, and the system-centric modeling perspective tends to overlook the agent’s perception, decision-making, and action processes. As a result, many existing digital twins remain in the static replica stage, lacking dynamic evolution and intelligent decision-making capabilities~\cite{Dagnaw2026digitaltwinsSuv}.

To complement and enhance the capabilities of digital twins, world models have gradually emerged and been adopted for similar tasks~\cite{zhao2025cognitive}. Their objective shifts from high-fidelity replication of the physical world to capturing environment evolution that is relevant to decision-making. This reflects a broader transition from world replication to task-relevant abstraction, from rule-driven to data-driven dynamics, and from a system-centric to an agent-centric modeling perspective.

World models typically use modern representation learning methods, such as variational autoencoders, to compress high-dimensional sensory input into low-dimensional latent states~\cite{ding2025understanding}. Action-conditioned state transitions are then learned in the latent space to support imaginative prediction of future environment dynamics. This paradigm aligns well with the core requirements of EGI, including resource efficiency, imagination-based planning, and close integration with reinforcement learning and control algorithms. Existing studies have explored world model design from multiple perspectives, including multi-scale variation modeling, controllable prediction, structured reasoning, and dynamics modeling~\cite{goff2025learning}. For example, DriveDreamer-2 generates diverse predictions covering long-tail scenarios in autonomous driving~\cite{Zhao2025DriveDreamer2}. GLAM improves model-based reinforcement learning by jointly modeling global and local state variations~\cite{He2025glamWorldModel}. Drive-OccWorld applies a vision-centric 4D world model to end-to-end autonomous driving planning~\cite{Yang2025visionCentric4dWorldModel}. SWAP uses world-model-driven entailment graphs for structured reasoning~\cite{Xiong2025swapWorldModel}\cite{Jia2026Sate}, and MoSim supports long-horizon physical state prediction through motion dynamics modeling~\cite{Hao2025MoSimWorldModel}. Thus, these studies show that by modeling only the environment dynamics related to task objectives and perceptual capabilities, world models can provide key support for long-term autonomy in resource-constrained edge environments.

\begin{figure*}[htbp!]
    \centering
    \includegraphics[width=0.85\textwidth]{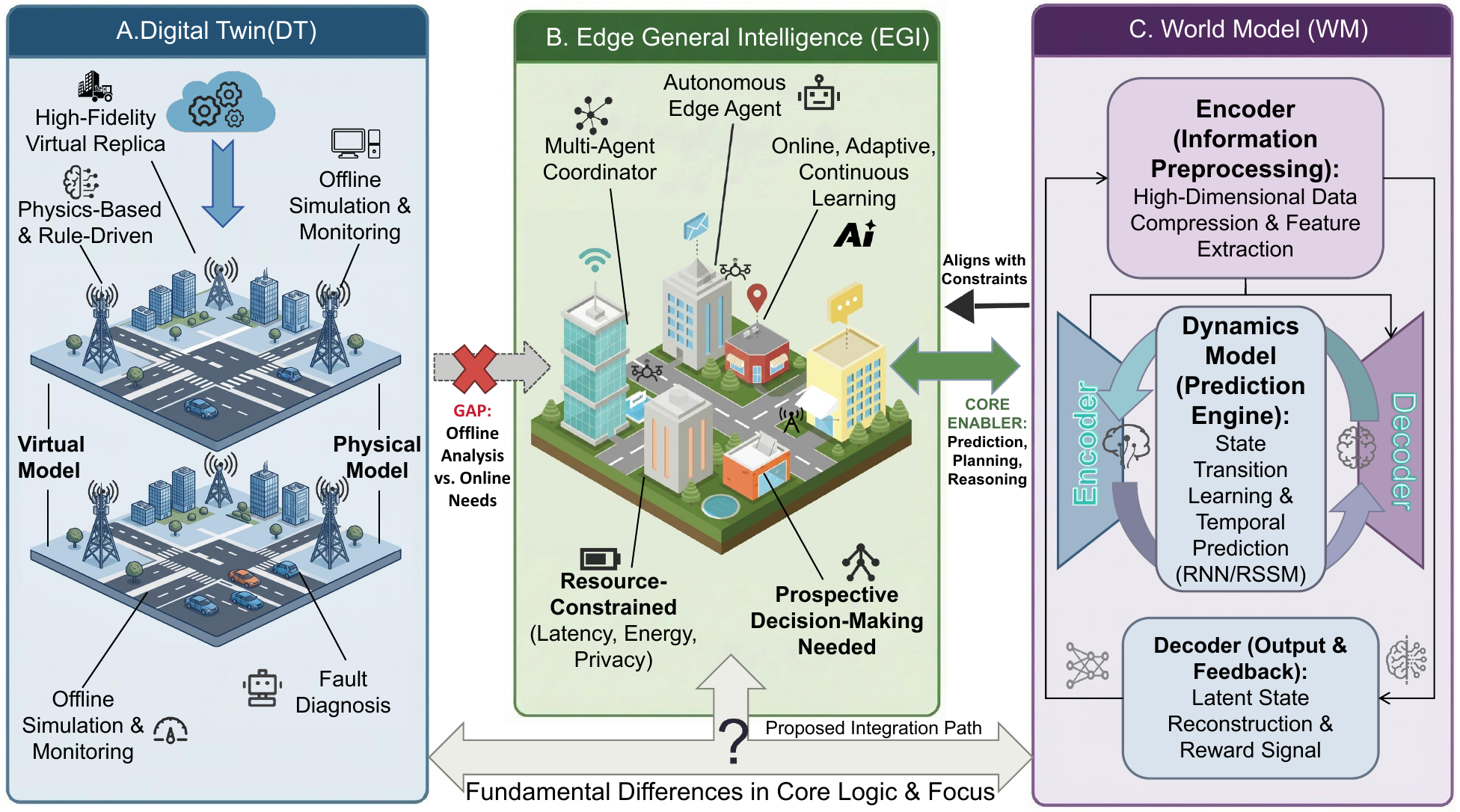}  % 从0.95改为0.85
    \caption{A conceptual framework illustrating how digital twins and world models enable EGI. (A): Digital twin offline physics replica feeds online monitoring. (B): EGI agents compress data and pick actions via reward. (C): world model hierarchy bridges offline planning to online action.}
    \label{fig:framework}
\end{figure*}

This survey reviews the evolution from digital twins to world models through the lens of EGI. Figure~\ref{fig:architecture} outlines the structure of this paper, covering motivation, world models, comparison with digital twins, technical evolution, applications, open resources, and future directions. Building on the background and motivation discussed above, the contributions of this work are as follows:
\begin{itemize}
\item We provide the first comprehensive review on world models for EGI in wireless and edge systems, and contrast them with traditional digital-twin-centric designs. We clearly distinguish world-model-centric EGI from conventional digital twin approaches, and outline a new direction for decision-centric modeling and autonomy at the network edge.  

\item We provide a systematic conceptual comparison and a unifying perspective on digital twins and world models as two complementary, yet fundamentally different, paradigms for modeling the physical world in edge systems. We highlight key shifts from world replication to decision-oriented abstraction, from rule-driven to data-driven dynamics, and from system-centric to agent-centric modeling, and explain how these shifts effectively match the requirements of EGI.  

\item We establish a taxonomy of world models tailored to wireless and edge scenarios, and decompose them into core components, including representation learning, dynamics modeling, observation and action interfaces, and imagination-based planning. We review representative methods from machine learning, robotics, and control, and reinterpret them in terms of edge deployment constraints, communication awareness, and integration with existing digital twin infrastructures.  

\item We demonstrate the potential of world models in edge systems by mapping classical digital twin applications in integrated sensing, communication, and computing (ISCC), semantic communication, air-to-ground network, and low-altitude platforms, and industrial edge infrastructures to their corresponding world model-based counterparts. We identify key open challenges and outline future research directions, including hybrid physics–data-driven world models, federated world modeling at the edge, multi-agent and multi-scale modeling, and explainable and trustworthy world models for safety-critical EGI.  
\end{itemize}

By organizing existing knowledge and open questions along these dimensions, this survey offers both a conceptual foundation and a practical roadmap for researchers and practitioners moving from digital-twin-centric design toward world-model-centric EGI.

\section{WORLD MODELS}
\label{sec:world_models}

This section introduces the basic concepts of digital twins and world models, and then discusses their main differences. We further explain how world models can serve as a core enabler for EGI. The detailed framework is shown in Figure~\ref{fig:framework}. 

\subsection{Introduction to Digital Twins}

Digital twin technology has emerged as a dominant paradigm for modeling physical systems in cyber-physical environments~\cite{Empl2025digitaltwinsSuv}. A digital twin is typically a high-fidelity virtual replica of a physical asset, system, or process governed by physical laws, domain-specific equations, and expert-crafted rules. Digital twins support detailed simulation, performance evaluation, fault diagnosis, and system optimization.

In wireless and edge systems, digital twins have been widely adopted for tasks such as network planning, configuration validation, interference analysis, and what-if performance evaluation under known operational conditions~\cite{10234399}. For example, network operators use digital twins to explore different base-station deployment strategies, predict coverage and capacity, and test control policies before applying them to the live network. These applications show that explicit physics-based and rule-based models can provide valuable insights into complex edge environments~\cite{huang2025dtgenai}.

However, the characteristics that make digital twins powerful for offline engineering analysis do not directly meet the needs of online, autonomous EGI agents that must learn, adapt and act continuously in the field. This gap motivates a critical reassessment of how world modeling should be approached for EGI.

\subsection{Overview of World Models}
\label{subsec:worldmodel_def_core}

\begin{table*}[htbp!]
\centering
\caption{Differences between digital twins and world models}
\label{tab:differences}
\renewcommand{\arraystretch}{0.85}
\small                                     % 字号降一档
\scalebox{0.95}
{
\begin{tabular}{@{}>{\raggedright\arraybackslash}p{2.2cm}@{\hspace{0.2cm}}%
                >{\raggedright\arraybackslash}p{1.0cm}@{\hspace{0.0cm}}%
                >{\raggedright\arraybackslash}p{6.0cm}@{\hspace{0.4cm}}%
                >{\raggedright\arraybackslash}p{4.2cm}@{\hspace{0.2cm}}%
                >{\raggedright\arraybackslash}p{4.0cm}@{}}
\toprule
\textbf{Name} & \textbf{Ref} & \textbf{Description} & \textbf{Technical Foundation} & \textbf{Advantages} \\
\midrule

\raisebox{-2\baselineskip}[0pt][0pt]{\textbf{World models}} &
\begin{minipage}[t]{1.0cm}
  \vspace{0pt}\raggedright
  \cite{ding2025understanding}\\\cite{ha2018recurrent}\\\cite{EGI2025chen}\\\cite{hafner2025nature}
\end{minipage}
&
\begin{minipage}[t]{6.0cm}
  \vspace{0pt}\raggedright
  \linespread{0.8}\selectfont
  \begin{itemize}[leftmargin=*, nosep, topsep=0pt, partopsep=0pt, parsep=0pt, itemsep=0pt]
    \item Learns compact latent representations from multi modal sensory inputs
    \item Models temporal state transitions to capture environment physics
    \item Forecasts future latent states conditioned on planned  actions	
  \end{itemize}
\end{minipage}
&
\begin{minipage}[t]{4.5cm}
  \vspace{0pt}\raggedright
  \linespread{0.8}\selectfont
  \begin{itemize}[leftmargin=*, nosep, topsep=0pt, partopsep=0pt, parsep=0pt, itemsep=0pt]
    \item Representation learning
    \item Dynamics models
    \item Self-supervised learning
    \item Latent-state compression
  \end{itemize}
\end{minipage}
&
\begin{minipage}[t]{4.5cm}
  \vspace{0pt}\raggedright
  \linespread{0.8}\selectfont
  \begin{itemize}[leftmargin=*, nosep, topsep=0pt, partopsep=0pt, parsep=0pt, itemsep=0pt]
    \item Resource-efficient
    \item Self-adaptive
    \item Foresighted
    \item Task-centric
  \end{itemize}
\end{minipage} \\
\midrule

\raisebox{-2\baselineskip}[0pt][0pt]{\textbf{Digital twins}} &
\begin{minipage}[t]{1.0cm}
  \vspace{0pt}\raggedright
  \cite{Empl2025digitaltwinsSuv}\\\cite{Mihai2022DigitalTA}\\\cite{tao2025modeling}\\\cite{tao2025smartmanu}
\end{minipage}
&
\begin{minipage}[t]{6.0cm}
  \vspace{0pt}\raggedright
  \linespread{0.8}\selectfont
  \begin{itemize}[leftmargin=*, nosep, topsep=0pt, partopsep=0pt, parsep=0pt, itemsep=0pt]
    \item Constructs physics-based virtual replicas of physical assets
    \item Maintains bidirectional synchronization between physical and virtual states
    \item Propagates state changes via real-time data streaming
  \end{itemize}
\end{minipage}
&
\begin{minipage}[t]{4.5cm}
  \vspace{0pt}\raggedright
  \linespread{0.8}\selectfont
  \begin{itemize}[leftmargin=*, nosep, topsep=0pt, partopsep=0pt, parsep=0pt, itemsep=0pt]
    \item IoT sensory ingestion
    \item Bidirectional synchronization
    \item Physics-based simulation
    \item Edge-cloud computing

  \end{itemize}
\end{minipage}
&
\begin{minipage}[t]{4.5cm}
  \vspace{0pt}\raggedright
  \linespread{0.8}\selectfont
  \begin{itemize}[leftmargin=*, nosep, topsep=0pt, partopsep=0pt, parsep=0pt, itemsep=0pt]
    \item Cyber-physical fidelity
    \item Life cycle governance
    \item Anticipatory risk mitigation
    \item Holistic visualization
  \end{itemize}
\end{minipage} \\
\bottomrule

\end{tabular}
}
\end{table*}

World models are internal simulations of environmental dynamics constructed by intelligent systems~\cite{ha2018recurrent}. By learning from real-world data and implicit laws, they capture key dynamic properties of the environment, predict future states, and provide agents with the ability to understand, reason about, and plan for the physical world. This capability of learning and representing physical dynamics enables world models to excel in computer vision tasks such as video generation. For instance, by internalizing spatiotemporal relationships from driving data, world models can generate realistic 4D scenes where objects follow consistent physical trajectories and spatial layouts over time, demonstrating their potential as physical simulators for visual content creation~\cite{zhang2025drivedreamer4d}.

World models are not high-fidelity reproductions of the physical world. Instead, they focus on dynamic aspects that are directly related to agent behavior and build an internal cognitive framework for imaginative reasoning~\cite{ding2025understanding}. This abstraction avoids the computational cost of full-scale replication and makes world models well suited to resource-constrained scenarios such as EGI, where they can offer efficient decision support.

Data-driven dynamic modeling is central to the adaptability of world models. Unlike traditional models that depend on preset physical formulas or rule bases, world models automatically extract implicit environmental laws including physical regularities and temporal correlations via unsupervised or self-supervised learning from multi-modal interaction data, such as sensor measurements, action feedback, and environmental observations~\cite{zhao2025cognitive}. For example, in UAV flight scenarios, they can learn the mapping between actions and states from data that reflect channel variations and meteorological interference~\cite{zhao2025cognitive}. Even when facing previously unseen low-altitude weather conditions, they can still make reasonable predictions based on learned general laws, thereby overcoming the generalization limits of traditional models and flexibly adapting to the dynamics and uncertainty of edge environments.

Prospective decision-making and planning are the core objectives of world models~\cite{ding2025understanding}. Rather than responding passively, they follow a prediction–imagination–decision pipeline that gives agents forward-looking cognitive capabilities. World models can simulate multi-step future evolutions of the environment based on the current state and candidate actions, assess the benefits and risks of different action sequences, and optimize decision strategies to avoid short-sighted behavior. This property makes them well aligned with the latency, energy, and privacy constraints that define EGI in practice~\cite{EGI2025chen}, and provides the core support for long-term autonomous and adaptive decision-making at the edge. This adaptability comes from three key design characteristics of world models for edge deployment.

The efficiency of the world model comes from latent-space abstraction. World models extract decision-relevant information without reconstructing the full environment in high fidelity, thus significantly reducing computation, storage, and communication overhead, which is critical for edge devices~\cite{zhao2025cognitive}. The agent-centric nature focuses only on the environmental dynamics that are relevant to the observations and actions of the agent, forming a localized task-oriented cognitive model and avoiding the use of limited resources on redundant information~\cite{guan2024world}. The generative imagination capability uses generative AI techniques to infer the trajectories of scenarios that have not yet occurred, improving sample efficiency and further adapting to the practical requirements of edge scenarios~\cite{zhao2025cognitive}. These characteristics complement the data-driven nature and prospective decision-making objective of world models and together strengthen their applicability in edge environments.

Thus, world models learn compact representations of environmental dynamics from data and, through generative imagination and prospective reasoning, provide efficient, autonomous, and adaptive decision support for resource-constrained edge agents. They are becoming an indispensable cognitive pillar in the realization of EGI.

\begin{table*}[htbp!]
\caption{From digital twins to world models: Enabling edge general intelligence}
\label{tab:edge_general_intelligence}
\centering
\footnotesize 
\small
\begin{tabular}{@{}p{2.2cm}|p{1.0cm}|p{5.0cm}|p{5.5cm}@{}}
\hline
\textbf{Scope} & \textbf{Ref} & \textbf{Key insight} & \textbf{From digital twins to world models} \\
\hline
\multirow{10}{=}{Edge general intelligence} &      
 ~\cite{tao2025saei}& 
\begin{itemize}[leftmargin=*,nosep,topsep=0pt]
\item Semi-synchronous edge intelligence using digital twins
\item Virtual-physical interaction for real-time decisions
\end{itemize}
& Digital twins continuously supply boundary data, laying an online learning foundation for later integration with world models \\
\cline{2-4}
 & ~\cite{zhao2025cognitive} & 
\begin{itemize}[leftmargin=*,nosep,topsep=0pt]
\item Lightweight recurrent state-space model with fast reasoning
\item Real-time prediction for autonomous edge nodes
\end{itemize}
& World models give edge nodes real-time look-ahead reasoning ability, providing a decision brain for subsequent connection to digital twins \\
\cline{2-4}
& ~\cite{zhou2026dtai} & 
\begin{itemize}[leftmargin=*,nosep,topsep=0pt]
\item Compress LLM knowledge into world models
\item Digital twins calibrate for trustworthy edge AI
\end{itemize}
& World models compress and offload large model capabilities to the edge, while digital twins correct physical errors in real time, making edge general intelligence both efficient and trustworthy \\
\hline
\end{tabular}
\end{table*}

\subsection{Differences Between World Models and Digital Twins}

Both world models and digital twins aim to model environmental dynamics and support decision-making, but their core logic and application focus are fundamentally different.

A world model is an \emph{internal cognitive framework} of an agent. Rather than replicating reality, it distills laws of environmental dynamics that are relevant to agents' actions and decisions~\cite{hafner2025nature}. Without relying on preset physical formulas or rule bases, a world model learns implicit environmental laws in an unsupervised or self-supervised way from multi-modal interaction data and is inherently designed to provide efficient decision support across diverse tasks~\cite{hansen2024tdmpc2}. To fit resource-constrained settings in EGI, world models reduce information dimensionality via latent-space abstraction, greatly lowering computation, storage, and communication overhead and avoiding resource use on redundant information. With a focus on optimizing future decisions, they enable agents to perform autonomous,adaptive and long-horizon decision-making in dynamic and uncertain edge environments, and have already shown clear advantages in navigation and related tasks that require accurate perception and decision-making~\cite{bar2025navworld}.

By contrast, the core of a digital twin is to build a \emph{high-fidelity virtual mirror} of a physical entity, pursuing the accurate correspondence between the virtual and actual systems~\cite{tao2025modeling}. Its modeling process heavily depends on preset physical formulas, rule bases, and rich sensor data, and aims to restore the real system states through physics-based modeling and data fusion~\cite{tao2025smartmanu}. This makes digital twins very effective for intelligent asset management,. However, such high-fidelity requirements lead to massive data transmission, complex simulations, and large storage demands, impose strict requirements on hardware resources, and typically rely on cloud or high-performance servers. Digital twins reside in real-time monitoring of physical entities, reconstruction of historical processes, and diagnosis of potential faults, with an emphasis on knowing the current real-world states. Thus, Digital twins are more suitable for Industry 4.0 manufacturing, urban operation and maintenance, and other scenarios that need fine-grained management, rather than highly resource-constrained edge scenarios.

The essential difference is that digital twins emphasize restoring reality, making the states and changes of physical entities observable and knowable, whereas world models emphasize predicting the future, making agent decisions more forward-looking and near-optimal through law distillation and imaginative reasoning~\cite{hafner2025nature,zhou2026dtai}. This difference in core orientation allows world models to overcome the strong resource dependence of digital twins and better match the latency, energy, and privacy constraints of edge scenarios, while digital twins remain indispensable in scenarios that require high-fidelity reconstruction.

\subsection{World-Model-Enabled Edge General Intelligence}
\label{subsec:worldmodel_egi}

Through leveraging the four core capabilities, such as imagination, prediction, planning, and reasoning, world models meet the autonomous decision-making needs of EGI under resource constraints and dynamic uncertainty~\cite{he2025road}.

\begin{figure*}[t]
    \centering
    \includegraphics[width=0.9\textwidth]{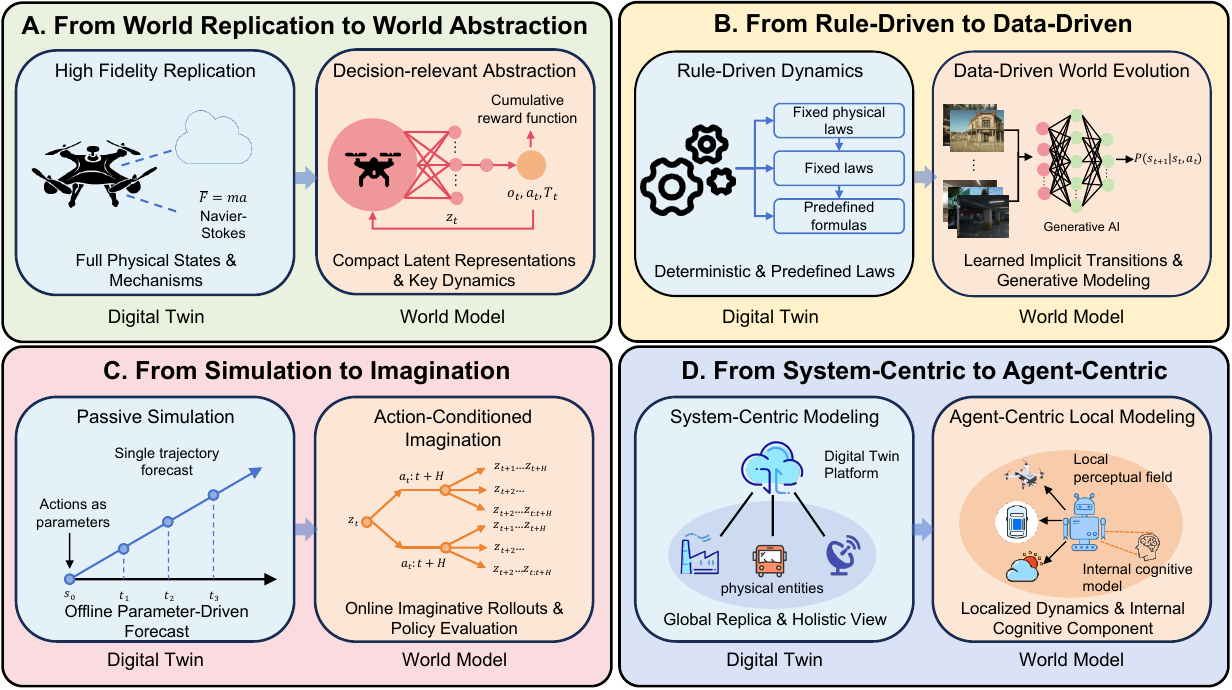}
    \caption{Evolution from Digital Twin to World Model for EGI.
    (A) From world replication to world abstraction, %: high-fidelity physical replication versus compact, decision-relevant latent representations.
    (B) From rule-driven to data-driven, %: predefined physical laws versus learned implicit state transitions and environment dynamics.
    (C) From passive simulation to active imagination, %: offline single-trajectory prediction versus action-conditioned multi-trajectory imagination and policy evaluation.
    (D) From system-centric to agent-centric.%: global system-level modeling versus agent-centric local perception and internal cognition.
    }
    \label{fig:DTtoWD}
\end{figure*}

\begin{itemize}

\item\emph{Imagination: }Imagination empowers world models to synthesize novel scenarios beyond observed reality, enabling agents to mentally explore hypothetical futures without physical interaction~\cite{ha2018recurrent}. This capability supports creative problem-solving and risk-free experimentation in resource-constrained edge environments. For example, by leveraging learned visual priors, navigation agents can imagine trajectories in unfamiliar environments from a single input image, predicting future visual observations to plan safe and efficient paths without prior exploration~\cite{zhang2025drivedreamer4d}.

\item \emph{Prediction:} Prediction refers to inferring future environmental states by learning implicit laws from multi-modal interaction data~\cite{zhao2025cognitive}. In EGI scenarios, a world model can foresee channel-quality fluctuations based on historical channel data and device trajectories, providing time margins for mode switching and power adjustment, and thus helping to avoid link outages~\cite{He2025glamWorldModel}. In low-altitude network, it can fuse historical and real-time weather data to predict short-term changes in wind speed and direction, provide early warnings for UAV flight safety, and mitigate the timeliness limitations of traditional weather forecasts~\cite{zhang2025aerialworld}.

\item \emph{Planning:} Planning uses predicted outcomes to simulate multi-step scenario evolution, evaluate candidate actions, and optimize decisions~\cite{wu2025kd}. When UAVs perform delivery or inspection tasks, the world model can combine weather and channel predictions to plan flight trajectories that balance safety, energy efficiency, and throughput. In scenarios with multiple UAVs or clusters of edge nodes, it can dynamically allocate limited computing, storage, and communication resources and adjust allocations based on task load and remaining resources, preventing overload or waste and improving overall system efficiency~\cite{wu2025kd}. EGI requires autonomous decision-making in uncertain environments, and world models enhance this ability through latent-space simulation. For example, Zhang et al. ~\cite{Zhang2024generativeaIagents} proposes a generative AI and multi-agent policy optimization framework for satellite communication, achieving efficient modeling and adaptive optimization, thereby improving the planning capability of EGI.

\item \emph{Reasoning:} Reasoning allows the world model to use learned general laws to handle unknowns and uncertainties at the edge~\cite{susnjak2025chaos}. In vehicular communications, when sudden road works or traffic jams occur, the onboard edge nodes can infer appropriate communication routes and trajectory adjustments by combining real-time road conditions with link states, thus maintaining continuous data delivery and driving safety~\cite{tang2025missing}. In industrial edge-device maintenance, a world model can analyze abnormal sensor readings, distinguish between actual faults, interference, and false alarms, and provide accurate support for operations and maintenance decisions, reducing unnecessary production loss~\cite{susnjak2025chaos}.

\end{itemize}

EGI face challenges such as time-varying channels and dynamic node appearance caused by high-speed mobility. The prediction, planning and reasoning capabilities of world models help ensure link stability, resource optimization, and rapid response to emergencies, thus meeting the low-latency and high-reliability requirements of wireless EGI networks~\cite{li2025online}. In low-altitude cooperative UAV scenarios, these capabilities jointly support flight safety and mission efficiency, and enable adaptation to the complex conditions of edge environments~\cite{He2025glamWorldModel}.

\section{From Digital Twin to World Model}\label{DTtoWM}
This section first introduces the core components of world models and then discusses the key technical shifts that drive the evolution from digital twins to world models.

\subsection{Core Components of World Models}
\label{subsec:worldmodel_components}

The perception–prediction–decision of world models is based on the coordinated operation of three core modules: the encoder, the dynamics model, and the decoder. These modules are tightly coupled and together form an efficient closed-loop workflow~\cite{zhao2025cognitive}.

\begin{itemize}
\item \emph{Encoder:} The encoder is the front-end information processing module of the world model and is mainly responsible for compressing high-dimensional data and extracting key features~\cite{zhao2025cognitive}. It receives multi-modal, high-dimensional perceptual data such as images and radar signals, removes redundant details via feature extraction algorithms, and preserves core information relevant to agent decision-making. In this way, it transforms high-dimensional raw data into compact low-dimensional latent states~\cite{shen2026metaworld}. This processing not only greatly reduces the computational overhead of downstream modules, which is important for edge devices, but also provides high-quality input for subsequent structure learning and prediction.

\item \emph{Dynamics model:} The dynamics model is the core engine that allows a world model to learn environmental evolution laws and perform temporal prediction~\cite{ha2018recurrent}. Its main role is to learn the state transition function of the environment so that it can accurately predict the next latent state given the current latent state and the action of agent~\cite{worden2026aiworld}. Dynamics models can be broadly divided into deterministic models, such as RNN-based models, and stochastic models, such as the recurrent state-space
model(RSSM), which uses both deterministic and stochastic hidden variables, and can be selected according to scenario requirements. The dynamics model takes the encoder’s latent state as input, combines it with the agent’s action, and predicts multi-step latent state trajectories, thereby providing crucial trend information to support decision-making~\cite{Hao2025MoSimWorldModel}.

\item \emph{Decoder:} The decoder is the key output module in the closed loop, responsible for mapping the latent states predicted by the dynamics model back to observable quantities~\cite{ding2025understanding}. It can reconstruct future environmental observation frames and output reward signals. These outputs can be used to visually verify prediction quality and assess the validity of latent states, and they also serve as an important basis for policy evaluation, helping the system select better decision strategies. In addition, the decoder’s outputs can be used as feedback signals to update the encoder and the dynamics model, continuously improving feature extraction and state prediction accuracy. This closes the perception–prediction–decision–iteration loop and ensures the integrity and effectiveness of the entire pipeline~\cite{Zhao2025DriveDreamer2}.
\end{itemize}

Thus, the progression from digital twin to world model for EGI involves a transition from rule-driven, system-centric world replication to a data-driven, agent-centric abstraction capable of active imagination, as shown in Figure~\ref{fig:DTtoWD} and Table~\ref{tab:dt_vs_wm2}.

\begin{table*}[htbp!]
\centering
\caption{Comparison between digital twin and world model}
\label{tab:dt_vs_wm2}
\footnotesize 
\renewcommand{\arraystretch}{1.0} 
\setlength{\tabcolsep}{2pt} 

\begin{tabularx}{\textwidth}{
l
>{\raggedright\arraybackslash}X
>{\raggedright\arraybackslash}X
>{\raggedright\arraybackslash}X
>{\raggedright\arraybackslash}X
>{\raggedright\arraybackslash}X
}
\toprule
\multicolumn{1}{l}{\textbf{Feature}} 
& \textbf{Characteristics}
& \textbf{Replication to Abstraction} 
& \textbf{Rule-based to Data-driven} 
& \textbf{Passive to Active} 
& \textbf{System-centric to Agent-centric} \\
\midrule

% 核心修改1：Digital Twin 水平+垂直强制双居中
\multirow{5}{*}{\centering\textbf{Digital Twin}}
&
\begin{itemize}[leftmargin=*, nosep,label={}]
  \item World Replication~\cite{Mihai2022DigitalTA}
  \item Rule-driven~\cite{Liu2024WhenDT}
  \item Passive Simulation~\cite{Tao2024AdvancementsAC}
  \item System-centric~\cite{Li2024DigitalTS}
\end{itemize}
& \begin{itemize}[leftmargin=*, nosep]
  \item Achieve high-fidelity mapping that reflects precise physical reality
  \item Maintain structural isomorphism to ensure geometric consistency with objects
  \item Reproduce intricate physical details and geometry for total synchronization
\end{itemize}
& \begin{itemize}[leftmargin=*, nosep]
  \item Operate based on deterministic rules to ensure predictable outcomes
  \item Models are strictly governed by explicit physical laws and mechanisms
  \item Utilize complex mechanism equations to simulate exact physical behaviors
\end{itemize}
& \begin{itemize}[leftmargin=*, nosep]
  \item Run simulations based on fixed parameters to verify designs
  \item Simulations rely on specific initial states to predict results
  \item Treat agent actions as external inputs within a static system
\end{itemize}
& \begin{itemize}[leftmargin=*, nosep]
  \item Provide comprehensive global modeling for the entire system landscape
  \item Create a complete virtual replica encompassing all system components
  \item Ensure full coverage of complex system behaviors and interactions
\end{itemize} \\
\midrule

% 核心修改2：World Model 水平+垂直强制双居中
\multirow{6}{*}{\centering\textbf{World Model}}
& \begin{itemize}[leftmargin=*, nosep,label={}]
  \item World Abstraction~\cite{ha2018recurrent}
  \item Data-driven~\cite{hafner2025nature}
  \item Active Imagination~\cite{zhao2025cognitive}
  \item Agent-internal cognitive component~\cite{ding2025understanding}
\end{itemize}
& \begin{itemize}[leftmargin=*, nosep]
  \item Abstract complex environments into compact, task-oriented representations
  \item Retain only the critical information necessary for reward prediction
  \item Extract low-dimensional latent features to simplify environmental complexity
\end{itemize}
& \begin{itemize}[leftmargin=*, nosep]
  \item Learn versatile environment representations through advanced generative AI
  \item Evolve through continuous learning from massive environmental interaction data
  \item Capture implicit state transition dynamics without explicit physical formulas
\end{itemize}
& \begin{itemize}[leftmargin=*, nosep]
  \item Predict future environment states conditioned on specific agent actions
  \item Imagine virtual trajectories within latent space to evaluate plans
  \item Enable autonomous exploration through internal "what-if" active simulations
\end{itemize}
& \begin{itemize}[leftmargin=*, nosep]
  \item Focus on ego-centric local modeling relevant to the agent
  \item Dynamically adapt modeling to match the agent's real-time perception
  \item Function as the internal cognitive brain for autonomous decision-making
\end{itemize} \\
\bottomrule
\end{tabularx}
\end{table*}

\subsection{From World Replication to World Abstraction}

The core principle of digital twins is world replication. The goal is to construct virtual systems that are highly consistent with the physical world in structure, state, and mechanisms, to support system-level simulation and monitoring~\cite{Mihai2022DigitalTA}.

The modeling objective of world models is world abstraction. Instead of pursuing full physical fidelity, world models retain only those environmental dynamics that affect an agent’s future cumulative rewards and emphasize decision-relevant information through abstract representations. For example, the world models framework proposed~\cite{ha2018recurrent} uses a variational auto-encoder(VAE) to learn low-dimensional latent representations and shows that such latent states alone can support complex control tasks with performance close to that achieved via direct interaction with the real environment. The Dreamer family proposed~\cite{hafner2025nature} further demonstrates that imaginative rollouts in a learned world model can significantly improve the sample efficiency of policy optimization.

When the modeling objective shifts toward online decision-making and control for EGI, the criteria for evaluating state representations differ fundamentally from those in traditional systems engineering. The objective of EGI is to bring general, adaptive, and context-sensitive cognitive capabilities to resource-constrained edge devices. It emphasizes long-term autonomous decision-making in dynamic, uncertain, and partially observable environments, under tight computation, storage, and communication budgets~\cite{Zhang2025TowardEG}.

In EGI scenarios, maintaining a globally accurate, real-time synchronized replica of the physical world leads to heavy computational and communication overhead and introduces large amounts of information that are irrelevant to current decisions, due to over-detailed physical descriptions~\cite{zhao2025cognitive}. This mismatch reduces the efficiency of decisions per unit of resource. Furthermore, traditional digital twins rely heavily on prior physical models, which are difficult to maintain in complex and unknown edge environments and often lack sufficient generalization and adaptability. Existing studies indicate that agent decision-making depends on how the environment evolves under the influence of actions rather than on complete reconstruction of physical states, while full replication of the physical world therefore does not produce an optimal state representation for decision-making~\cite{Meuser2024RevisitingEA}.

Abstraction-based world models show clear advantages for EGI. The core requirement of EGI is to capture the mapping between scheduling or control actions and system performance, rather than to reconstruct the full physical state. By discarding irrelevant details for decision-making, world models can significantly reduce computation and communication costs on the edge~\cite{zheng2025world4driveendtoendautonomousdriving}. More importantly, by learning environmental dynamics directly from interaction data, world models reduce dependence on precise prior physical models and lay the foundation for strong generalization and long-term autonomy in unknown environments~\cite{long2025survey,Fang2024TowardsUA}.

The evolution from digital twins to world models thus reflects a decision-centered shift in design philosophy. The modeling goal changes from building a high-fidelity replica of the world to constructing an abstract world representation that supports efficient prediction, planning, and learning by agents.

\subsection{From Rule-Driven Dynamics to Data-Driven World Evolution}

Traditional digital twin technology is grounded in physics-based mechanistic models, in which environmental evolution is governed by predefined physical laws and system-level equations. This rule-driven paradigm provides deterministic descriptions of physical entities through high-fidelity virtual replicas~\cite{Mihai2022DigitalTA}.

World models adopt a data-driven modeling paradigm. Using generative AI techniques, they compress environmental dynamics into compact latent spaces. By analyzing interaction data between agents and their environments, world models can autonomously learn implicit laws of state transitions and predict future environmental states~\cite{ha2018recurrent}.

A key characteristic of EGI is the need for long-horizon autonomous decision-making in dynamic, unstructured, and highly uncertain physical environments~\cite{Deng2019EdgeIT}. In practical EGI scenarios, such as UAV networks and intelligent transportation systems~\cite{9866918}, fixed rule-based models often fail to capture the long-term evolution of complex systems with sufficient accuracy. This rule-driven nature limits their generalization capability. Since a digital twin operates strictly according to predefined formulas and assumptions, it cannot flexibly adjust its evolution logic when facing unmodeled physical phenomena or unexpected disturbances beyond its design scope. In addition, the complex numerical solvers required for high-fidelity physical modeling impose a substantial computational load that conflicts with the limited resources of edge devices.

Data-driven world evolution modeling becomes a key direction for EGI. EGI agents can continuously refine their internal world models using local interaction data, thus adapting to environmental changes without redesigning physical equations~\cite{zhao2025cognitive}. By performing evolution inference in low-dimensional latent spaces, world models avoid pixel-level simulation of high-dimensional physical entities and significantly reduce computational latency and resource consumption. For example, in UAV networks, the aerial network world model, trained in large-scale aerial sequences, trajectories, and semantic labels, can predict semantically plausible long-range scenarios even under unseen meteorological conditions. This supports the dynamic generation of navigation paths that balance obstacle avoidance and high-level semantic objectives~\cite{zhang2025aerialworld}.

The shift from rule-driven to data-driven modeling represents a fundamental evolution of EGI rather than a simple technical improvement. This transition allows the system to move beyond passive reproduction toward proactive prediction within open-world environments

\subsection{Action-Conditioned Prediction and Imagination}

As EGI evolves from basic perception to complex autonomous decision-making, the objective of modeling shifts from describing current states to reasoning about the consequences of actions~\cite{Ali2025EdgeAI}. From this viewpoint, digital twins and world models represent two distinct predictive paradigms: the former focuses on parameter-driven offline simulation, whereas the latter supports action-conditioned online imagination.

Traditional digital twins are designed mainly for system monitoring and performance evaluation. Their predictive mechanisms are typically based on passive simulation of environmental dynamics. Given an initial state and boundary conditions, a digital twin carries out deterministic simulations to forecast system trajectories. Agent actions are often treated as external input or configuration parameters, rather than as intrinsic drivers of system evolution~\cite{He2025IntegratingIA}.

By contrast,world models use an action-conditioned prediction mechanism that explicitly models the agent’s actions as key variables in state transitions. Since trial-and-error interaction in the physical world is often costly, world models allow agents to conduct imaginative trajectory rollouts in a compact latent space. Without directly interacting with the real environment, an agent can rapidly simulate multiple candidate action sequences inside the model and evaluate their cumulative long-term rewards~\cite{ding2025understanding}.

In EGI scenarios, edge devices are constrained by limited computing, storage, and communication resources, while EGI tasks often involve long time horizons and limited rewards~\cite{Liu2025IntegratedSA}. Agents must therefore compare many potential decision paths under stringent cost constraints, which is difficult to achieve efficiently with prediction schemes based on high-fidelity simulation.

World models perform action-conditioned prediction and imagination in low-dimensional latent spaces. This avoids pixel-level simulation of high-dimensional physical states, greatly reducing computational complexity while preserving essential dynamics for decision-making. In autonomous driving, for example, FSDrive~\cite{zeng2025futuresightdrive} employs a world model as a predictor to generate future imagined scenes in latent space that jointly capture spatial and temporal structure, thereby enabling trajectory planning and policy evaluation without explicit pixel-level physical simulation.

By shifting prediction from system-level reproduction to action-driven evaluation of future outcomes, world models provide a more efficient and scalable decision-support paradigm for edge intelligence.

\subsection{Agent-Centric and Localized World Modeling}

Digital twin technology originated in large-scale system engineering contexts such as Industry 4.0 and smart cities, and its design philosophy is inherently system-centric~\cite{Tao2024AdvancementsAC}. By constructing a global virtual replica of the entire physical system, digital twins aim to reproduce the states of physical entities, their structural relationships, and their evolution from a holistic perspective, thereby supporting system-level analysis, simulation, and management.

World models adopt an agent-centric and localized modeling paradigm. Agent-centric modeling means that a world model does not try to reconstruct the full objective physical world. Instead, it serves as an internal cognitive component of the agent~\cite{Baraldi2025TheSC}. The model is defined by the agent’s sensing capabilities, action space, and task objectives, and captures only the environmental dynamics that are relevant to the agent’s observations and actions. Through localized modeling, world models learn implicit evolution patterns and retain only local dynamics that directly affect current decisions~\cite{Li2025ACS}. Thus, the size of the model is decoupled from the overall environmental complexity, allowing resource-constrained edge devices to perform complex environment modeling tasks.

In EGI scenarios that require continuous autonomous decision-making and online adaptation, the primary challenge of an agent is not only to precisely reconstruct the environment, but also to quickly understand local conditions, predict the consequences of actions, and make effective decisions under partial observability~\cite{Luo2025TowardEG}. For edge-deployed agents, sensing range, communication capability, and computational resources are limited, and the global state of the environment is generally neither fully observable nor practically maintainable~\cite{Zhang2025TowardEG}.

The agent-centric and localized nature of world models closely matches the fundamental requirements of EGI~\cite{Wang2025OptimizingEA}. This modeling paradigm reduces the complexity of model construction and maintenance, making online learning and continuous updates on edge devices more feasible. It supports a deep integration with decision-making methods such as reinforcement learning and model predictive control, allowing the world model to function as an internal part of the agent’s decision process rather than an external simulation tool. Localized world models offer better scalability and robustness in typical EGI scenarios characterized by multi-agent operation, non-stationary environments, and incomplete information\cite{Yang2025WorldRFTLW}.

By shifting from system-centric to agent-centric modeling, world models focus on local dynamics directly relevant to individual decisions. This shift not only reduces resource consumption at the edge,but also provides essential internal support for efficient prediction, reasoning, and autonomous decision-making in dynamic physical environments.

\subsection{Summary of the World Model Paradigm}

This section has examined the paradigm shift from digital twins to world models. This transition redefines physical-world modeling from fidelity-oriented replication to decision-oriented abstraction, establishing a more suitable cognitive modeling framework for resource-constrained EGI. Therefore, we summarize the key characteristics of the world model paradigm in four dimensions.
\begin{itemize}
    \item \emph{Decision-Oriented Abstract Modeling:}
With high-fidelity physical replication of digital twins, world models abstract the environment in a decision-oriented manner by learning compact latent representations that preserve only the dynamics relevant to an agent’s future rewards. This abstraction reduces model dimensionality and computational overhead, making world models well suited for resource-constrained edge devices~\cite{ha2018recurrent}.

\item \emph{Data-Driven Evolutionary Learning:}
Rule-driven digital twins that rely on predefined physical equations, world models use data-driven generative modeling to learn state transition dynamics from agent–environment interactions. This approach captures complex dynamics that are difficult to describe analytically and supports continual learning, enabling the model to adapt to environmental changes and unexpected disturbances~\cite{zhao2025cognitive}.

\item \emph{Action-Conditioned Imaginative Reasoning:}
World models incorporate agent actions into state transitions through action-conditioned prediction. By iterating one-step predictions, they enable imagination-based reasoning in latent space to evaluate future trajectories under different actions. This allows agents to assess long-term effects without real-world interaction, supporting forward planning and improving sample efficiency~\cite{Fang2024TowardsUA}.

\item \emph{Agent-Centric Local Modeling:}
As an internal cognitive component, a world model is shaped by the agent’s perception, action space, and task goals, and captures only the local dynamics the agent can observe and influence. This agent-centric design decouples model complexity from the global environment, reduces the reliance on global state synchronization, and supports heterogeneous and collaborative deployment in distributed edge systems~\cite{ding2025understanding}.
\end{itemize}
%These four characteristics are closely related and jointly define the basic paradigm of world models. Abstract modeling and data-driven learning make it possible to efficiently acquire environmental regularities; action-conditioned prediction turns the environment model from a passive observation tool into a cognitive basis for active decision-making; and agent-centric localization embeds the model inside the agent’s autonomy rather than treating it as an external, standalone system.

In EGI, the world model paradigm provides a coherent and practical framework for environmental modeling. With replacing digital twins in system-level analysis, world models complement them by supporting internal cognition and decision-making of agents. This enables edge agents to achieve continuous autonomy and long-term adaptation in dynamic, partially observable, and resource-constrained environments, marking a shift from task-specific automation to more general autonomous intelligence.

\section{Applications of From Digital Twins to World Models}\label{app}
\begin{figure*}[t]
    \centering
    \includegraphics[width=0.9\textwidth]{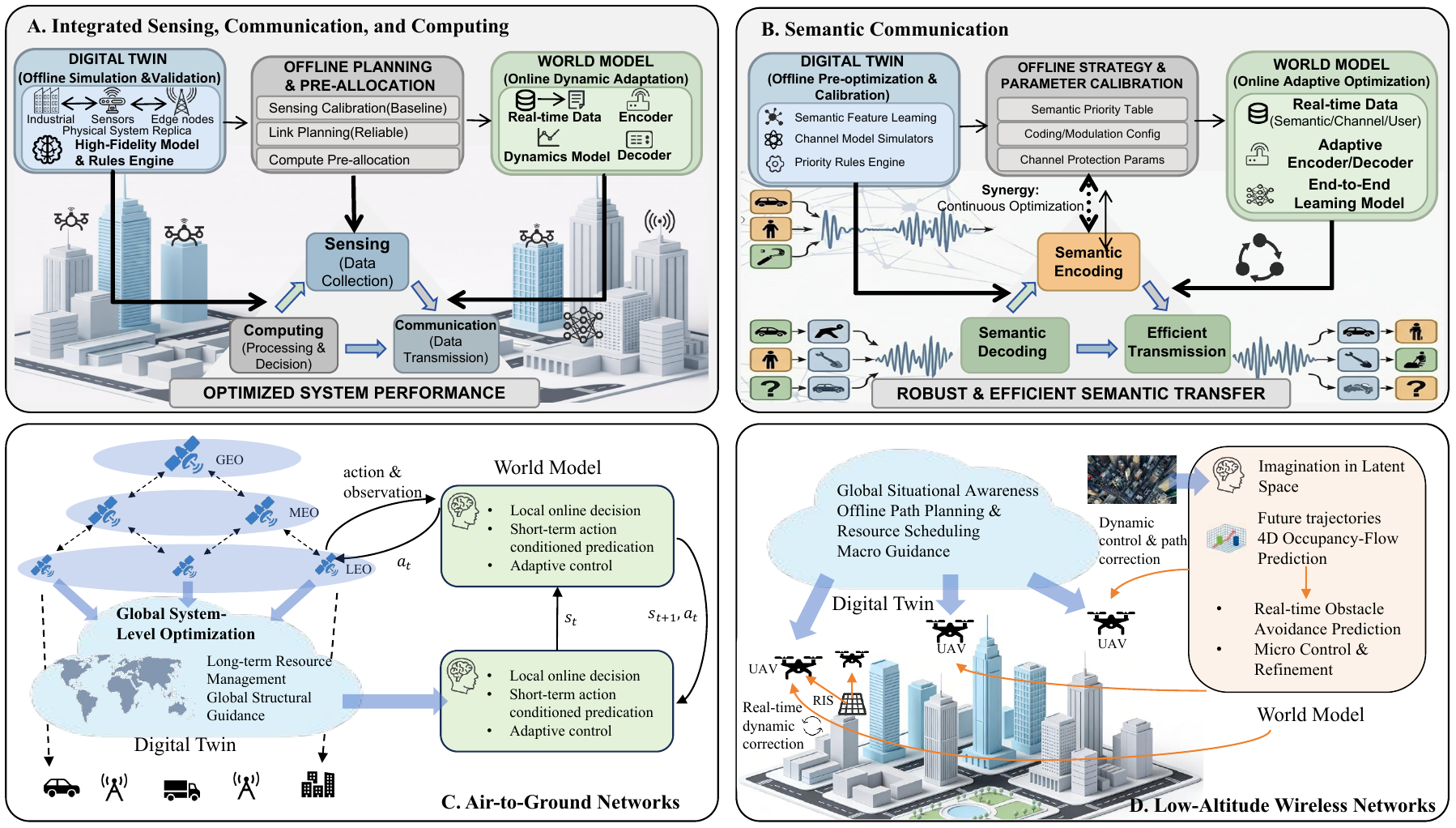}
    \caption{Evolution from digital twin to world model for EGI.
    (A) From world replication to world abstraction, %: high-fidelity physical replication versus compact, decision-relevant latent representations.
    (B) From rule-driven to data-driven, %: predefined physical laws versus learned implicit state transitions and environment dynamics.
    (C) From passive simulation to active imagination, %: offline single-trajectory prediction versus action-conditioned multi-trajectory imagination and policy evaluation.
    (D) From system-centric to agent-centric.%: global system-level modeling versus agent-centric local perception and internal cognition.
    }
    \label{fig:application}
\end{figure*}

This section discusses the application of world models to integrated sensing, communication, and computing
(ISCC), semantic communication, air-to-ground networks, and low-altitude wireless networks. Table~\ref{tab:DT_WM_comparison} compares digital twins and world models based on their main functions and related studies in these scenarios.

\subsection{Integrated Sensing, Communication, and Computing}
\label{subsec:iscc}

\begin{figure}[htbp]
    \centering
    \includegraphics[width=\columnwidth]{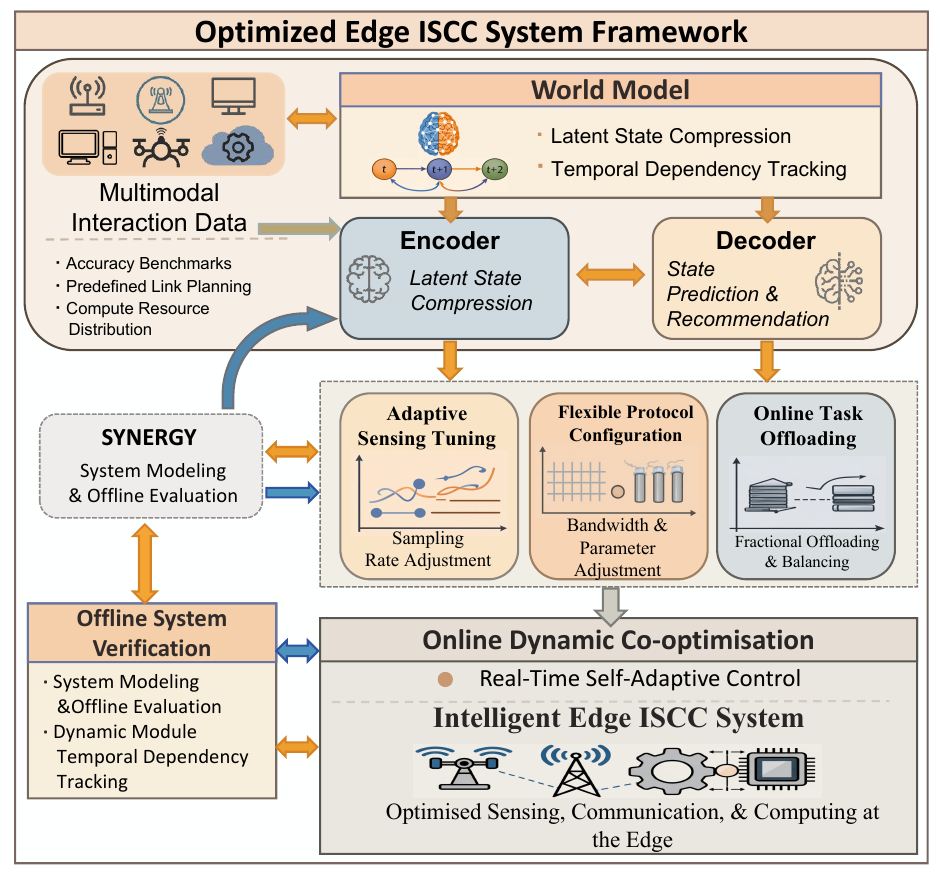}
    \caption{A conceptual framework illustrating how the optimized edge ISCC system enables intelligent edge computing and sensing through multi-modal data processing, world model-driven perception, and dynamic edge co-optimization.}
    \label{fig:5}
\end{figure}

Integrated sensing, communication, and computation (ISCC) is a key application in edge intelligence. Its objective is to orchestrate the end-to-end optimization of these resources under strict edge constraints, thereby maximizing system performance in connectivity-centric scenarios~\cite{wen2025survey}. Digital twins and world models jointly support deep synergy among the three ISCC pillars. The digital twin provides offline verification for sensing-data calibration, communication-link planning and computational-resource pre-allocation, whereas the world model dynamically adjusts, in an online fashion, sensing frequency, communication-protocol parameters and computation-offloading policies. The two mechanisms are functionally complementary and together form a cohesive co-design paradigm.
The two mechanisms are functionally complementary and together form a cohesive co-design paradigm, as illustrated in Fig.~\ref{fig:5}.

The principal value of the digital twin lies in its ability to perform offline, system-level modeling and validation of sensing accuracy, link quality, and resource allocation~\cite{li2023adaptive}. By constructing a high-fidelity virtual replica of the physical system, the twin exploits predefined physical models and rules to run simulations. In industrial-edge deployment scenarios, it accurately reproduces sensor coverage, wireless signal attenuation, and compute-node capacity distributions, enabling ex-ante evaluation of various resource-scheduling policies~\cite{li2025budget}. This high-fidelity emulation offers reliable evidence for offline resource allocation and task scheduling, eliminates the risks and overhead of direct experimentation on real hardware, and constitutes an indispensable tool during the ISCC design phase.

The world model focuses on online, dynamic co-optimization of sensing, communication, and computation, thereby accommodating the volatility and uncertainty inherent to edge environments. Its core advantage is prospective planning and real-time adaptation conditioned on the instantaneous system's state \cite{delser2025world}. The world model employs data-driven approaches to learn the latent dependencies between the three functions using multi-modal interaction data~\cite{chen2025distributed}. To achieve this, an encoder first compresses high-dimensional data into low-dimensional latent states, extracting critical information to reduce overhead. Next, a dynamics module tracks environmental changes and captures temporal dependencies. Finally, a decoder maps these predicted states back to observable quantities, completing the sense-predict-decide loop. In ISCC scenarios for UAV swarms, the model forecasts communication traffic and computational-resource variations induced by alternative trajectory adjustments, enabling proactive cooperative strategies~\cite{ma2025uav}. In industrial-edge contexts, it adaptively tunes sensor sampling rates, communication protocol configurations, and offloading fractions, dynamically balancing competing objectives~\cite{deng2024adaptive}. This online adaptability exactly compensates for the inability of offline digital-twin optimization to cope with environmental dynamics. In physical layer security, the prediction and generation abilities of world models have shown great potential. The APEG~\cite{cheng2026apeg} framework uses generative AI and diffusion models to achieve high-accuracy and adaptive physical layer authentication in dynamic environments.

The synergy between digital twins and world models propels ISCC from offline static optimization to online dynamic self-adaptive optimization~\cite{zhao2025cognitive}. Exploiting high-fidelity physical mapping and offline deduction, the digital twin delivers precise calibration benchmarks for sensing, dependable link-planning evidence for communication, and rational resource pre-allocation schemes for computation. Leveraging data-driven predictive modeling and online decision making, the world model continuously refines sensing frequency, communication parameter configurations and computation-offloading strategies at run time~\cite{wen2026federated}, achieving online co-alignment of sensing, communication and computation. Collectively, the integration of digital twins and world models breaks down silos-based subsystem optimization, endowing ISCC with both the reliability of offline planning and the agility of online adaptation. Specifically, digital twin-integrated frameworks achieve the 33\% end-to-end latency reduction and a 30\% reduction in tail latency at a mere 2--7\% energy cost~\cite{li2023adaptive}, while world models enable dynamic parameter adjustment of 15--20\% lower execution latency~\cite{deng2024adaptive}, ultimately enhancing system stability and efficiency in complex edge scenarios.

\begin{table*}[t]
\centering
\footnotesize
\caption{Functional comparison of digital twin and world model in edge intelligence}
\label{tab:DT_WM_comparison}
\renewcommand{\arraystretch}{1.15}
\setlength{\tabcolsep}{4pt}

\begin{tabular}{
>{\centering\arraybackslash}m{2.5cm}
>{\centering\arraybackslash}m{1.5cm}
>{\centering\arraybackslash}m{2.5cm}
m{6.0cm}
}

\toprule
\textbf{Scenario} & \textbf{Paradigm} & \textbf{References} & \textbf{Key Functions and Contributions} \\
\midrule

\multirow{2}{*}{ISCC} 
& Digital Twin 
& \cite{li2023adaptive,li2025budget,zhao2025cognitive} 
& Supports high-fidelity simulation for offline sensing calibration, link configuration, and resource pre-allocation. \\
\cmidrule(lr){2-4}
& World Model 
& \cite{delser2025world,chen2025distributed,ma2025uav,deng2024adaptive} 
& Enables online coordination via latent dynamics learning and adaptive sampling and offloading. \\

\midrule

\multirow{2}{2.5cm}{\centering Semantic Communication} 
& Digital Twin 
& \cite{li2025delay,okegbile2026novel,tang2025semantic} 
& Provides semantic policy calibration through priority setting and channel loss simulation. \\
\cmidrule(lr){2-4}
& World Model 
& \cite{jiang2025semantic,tan2025scenediffuser,yang2025fluid} 
& Supports dynamic semantic adaptation through implicit channel learning and encoding adjustment. \\

\midrule

\multirow{2}{*}{A2G Networks} 
& Digital Twin 
& \cite{Hevesli2024TaskOO,Gong2025BlockchainAidedDT,Lin2025AIDrivenSA} 
& Facilitates global multi-layer scheduling and cross-layer system optimization. \\
\cmidrule(lr){2-4}
& World Model 
& \cite{zhang2025aerialworld,Lu2025RemoteSW} 
& Enables low-latency local control via action--link modeling and latent roll out. \\

\midrule

\multirow{2}{*}{LAWNs} 
& Digital Twin 
& \cite{Xie2023RadarIB,Wang2025ComputingPI} 
& Supports 3D environment modeling and trajectory optimization for coverage planning. \\
\cmidrule(lr){2-4}
& World Model 
& \cite{bar2025navworld,Diehl2025DIODI,Chen2025HybridNA} 
& Enables onboard real-time control with obstacle avoidance and channel prediction. \\

\bottomrule
\end{tabular}
\end{table*}

\subsection{Semantic Communication}
\label{sec:semantic}

\begin{figure}[htbp]
    \centering
    \includegraphics[width=0.95\columnwidth]{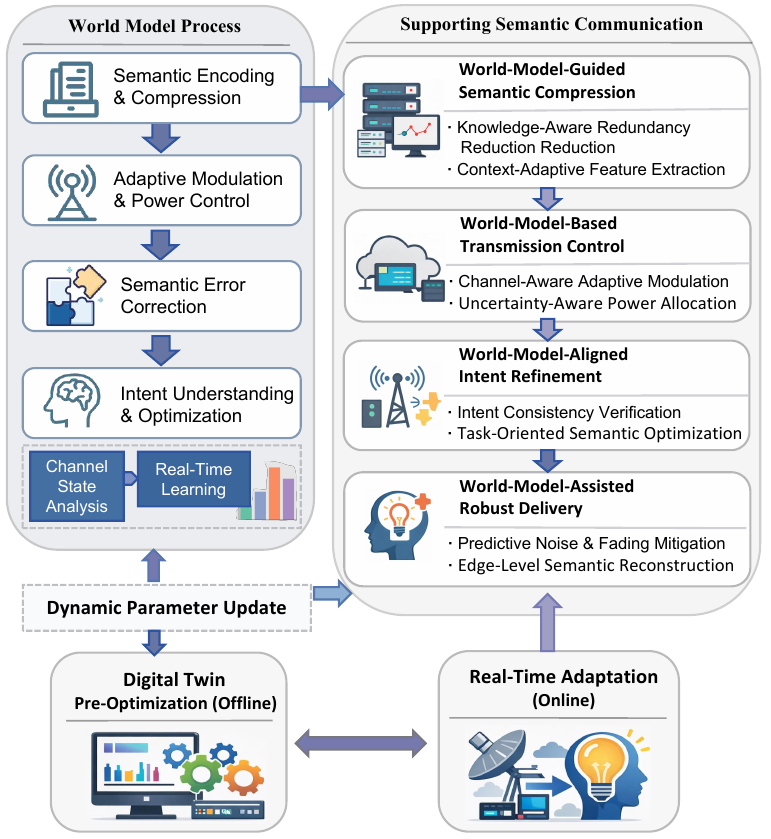}
    \caption{A conceptual framework depicting the integration of world model processing, digital twin pre-optimization, and real-time adaptation for enabling robust and intelligent semantic communication.}
    \label{fig:6}
\end{figure}

Semantic communication is widely recognized as a key technique for next-generation 6G networks because it focuses on the efficient transmission of semantic essence and core intent~\cite{chaccour2025less}. By conveying semantic meaning rather than raw bits, it breaks the throughput ceiling of conventional schemes and remains efficient under limited bandwidth or hostile channels, thus providing a key enabler for resource-constrained intelligent networking~\cite{Zheng2024Energy}. The digital twin is responsible for offline pre-optimization and parameter calibration of semantic policies, whereas the world model performs online adaptation by tracking end-to-end dynamics and continuously refining encoding, transmissions, and decoding. The two actors therefore operate in a complementary offline design, and an online evolution loop, as illustrated in Fig.~\ref{fig:6}.

The digital twin, operating within semantic communication, concentrates on noise-prone channels, priority-aware semantic priority and the stringent requirement for guaranteed recognition accuracy, thereby enabling offline pre-optimization and calibration of core transmission parameters~\cite{li2025delay}. By learning intrinsic semantic features and their interaction rules, the digital twin proactively simulates the suitability of candidate encoding or transmission schemes. The digital twin tests the semantic-loss rate of a specific code under complex noise and measures the delivery delay of critical intent messages, then refines semantic recognition and reconstruction algorithms accordingly. Priority assignment is calibrated against latency budgets~\cite{li2025delay}, while channel adaptation is tuned to counteract time-varying channel fading~\cite{okegbile2026novel}. This targeted offline optimization intrinsically improves reliability and efficiency, mitigates semantic distortion or intent loss, and lays a stable foundation for deployment, especially in scenarios where business demands are static and semantic rules are well defined~\cite{tang2025semantic}.

The world model focuses on the real-time capture of dynamic variations along the entire semantic link. Without assuming preset channel parameters, it autonomously learns the latent relationship between semantic content, instantaneous channel state and user demand from massive live data and performs end-to-end optimization~\cite{jiang2025semantic,tan2025scenediffuser}. The world model first compresses semantic data in the encoder to eliminate redundancy. It then adapts modulation and power during transmission to mitigate fading and, finally, corrects semantic distortion at the decoder to preserve the original intent. Its lightweight design enables its deployment on edge nodes, facilitating an efficient and robust cloud-edge-terminal architecture~\cite{yang2025fluid}.

In the offline phase, the digital twin generates semantic priority tables, code modulation settings, and channel protection parameters~\cite{tang2025semantic}. During online operation, the world model dynamically updates these parameters to adapt to real-time noise, traffic, and resource conditions. This interaction ensures that the end-to-end system maintains near-optimal performance automatically, thus delivering the intended information with stability and efficiency~\cite{jiang2025semantic}. Semantic communication empowered by world models provides a scalable and reliable architecture for next-generation intelligent networks.

\subsection{Air-to-Ground Networks}

\begin{figure}[htbp]
    \centering
    \includegraphics[width=0.95\columnwidth]{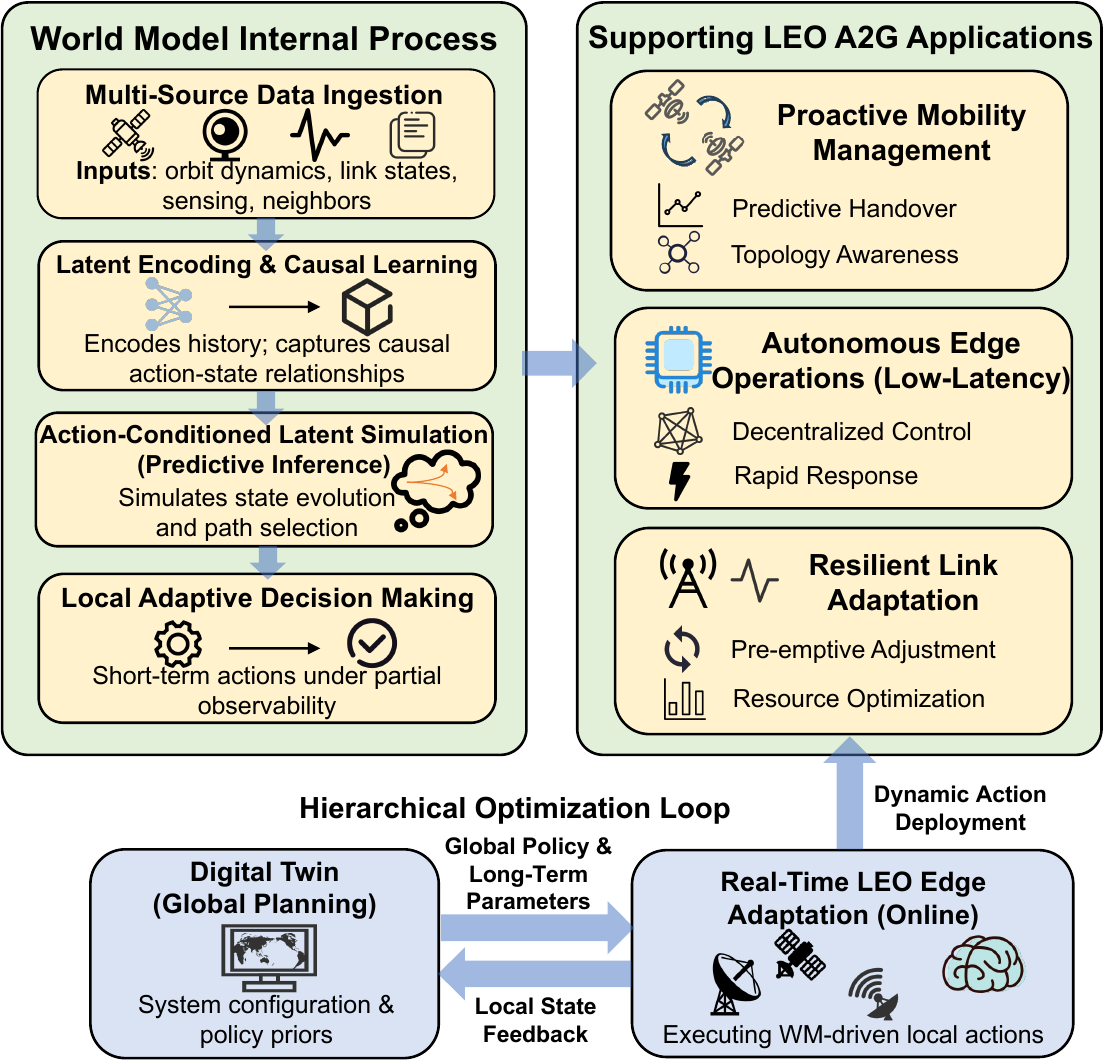}
    \caption{Overview of world model–enabled framework for LEO–A2G networks: Latent simulation and causal learning with digital twin–edge hierarchical optimization supporting mobility, low-latency operations, and link adaptation.}
    \label{fig:A2G}
\end{figure}

Air-to-Ground (A2G) networks integrate satellites, high-altitude platforms, and UAVs to form a three-dimensional space-air–ground (SAG) communication and computing architecture, providing fundamental support for wide-area connectivity, low-latency services, and edge intelligence. However, highly dynamic aerial topologies, heterogeneous network resources, and stringent ultra-low-latency requirements pose significant challenges to efficient resource allocation and task offloading~\cite{Hu2025GenerativeAS}. Digital twins focus on global modeling of the multi-layer air–space–ground network structure and resource constraints, enabling system-level optimization for cross-layer resource scheduling and task offloading. To illustrate the hierarchical interaction between global planning and local adaptive control in A2G networks, 
the integrated Digital Twin–World Model framework is depicted in Fig.~\ref{fig:A2G}.

Digital twins support system-level resource management and computation offloading in A2G scenarios by constructing high-fidelity virtual replicas of SAG networks. Hevesli et al.~\cite{Hevesli2024TaskOO} developed a digital twin-based framework for air–ground cooperation in the 6G industrial Internet of Things, where a virtual network replica is used for real-time state prediction. Gong et al.~\cite{Gong2025BlockchainAidedDT} proposed a SAG digital twin integrated blockchain architecture for air–space–ground heterogeneous networks, enabling centralized digital twins to perform global resource scheduling and ensure globally optimal task allocation.

As local environment simulators for A2G edge nodes, world models emphasize the relationships between aerial agent actions and local link and airspace states, supporting short-term prediction under partial observability. Zhang et al.~\cite{zhang2025aerialworld} proposed an aerial network world model that encodes historical link states, flight trajectories, and control actions into compact latent representations, allowing path selection to be simulated in latent space. Lu et al.~\cite{Lu2025RemoteSW} integrate edge interaction data with aerial and satellite sensing information, enabling world models to learn action-conditioned spatial state evolution and perform predictive inference before link fluctuations occur, thereby reducing the dependence of aerial nodes on frequent synchronization with centralized controllers. Zhang et al.~\cite{Zhang2025lmae} propose an integrated air–ground edge–cloud framework that improves multi-modal AI inference under limited bandwidth and outperforms both cloud-only and edge-only schemes.

In A2G networks, digital twins and world models exhibit clear functional separation. Digital twins act as global operational mappings and planning centers for A2G networks, addressing how the general system should be configured~\cite{Lin2025AIDrivenSA}. World models serve as local cognition and decision engines for aerial edge agents, determining which actions should be taken under current local airspace and link conditions. Together, they form a hierarchical optimization framework that spans from system-level planning to link-level adaptive control.

\subsection{Low-Altitude Wireless Networks}

\begin{figure}[htbp]
    \centering
    \includegraphics[width=0.95\columnwidth]{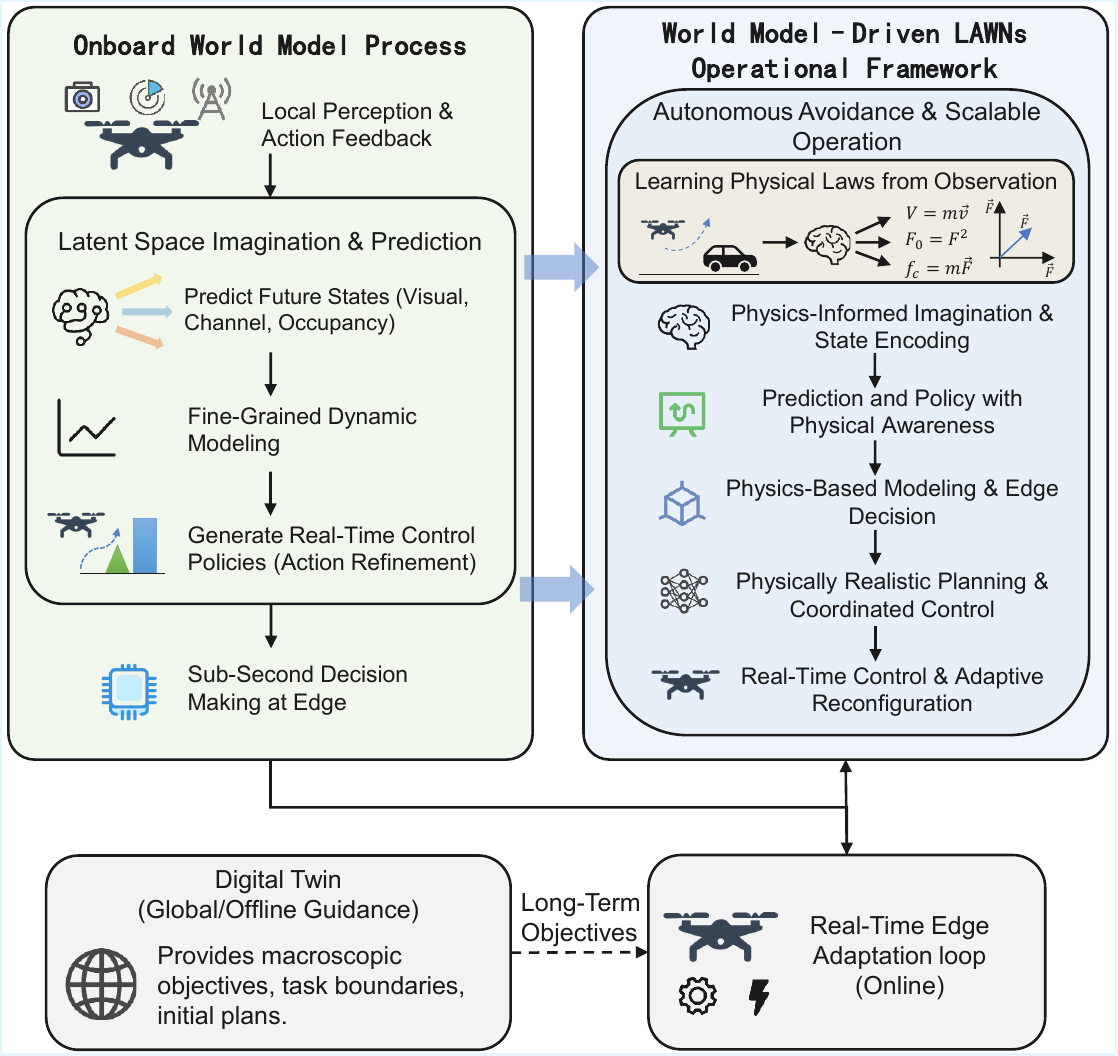}
    \caption{Overview of the world model–driven framework for LAWNs: Onboard latent prediction, physics-aware control, and digital twin–guided edge adaptation.}
    \label{fig:LAWNs}
\end{figure}

Low-Altitude wireless networks (LAWNs) are composed of platforms such as UAVs and electric vertical take-Off and landing (eVTOLs), typically operating below an altitude of 3000 m. In these networks, highly unstructured environments, strong uncertainty, and strict computation and energy constraints on board make global planning and fast local response equally important~\cite{Cai2025SecurePL}. Digital twins enable global awareness of aerial platform states by constructing high-fidelity virtual replicas of LAWNs and provide macroscopic guidance for large-scale resource scheduling. World models act as embodied cognition onboard aerial platforms. Using imagination in latent space, they empower UAVs with real-time local control capabilities, such as obstacle avoidance, short-term channel prediction, and action refinement in complex environments~\cite{Zhao2025AirScapeAA}. As shown in Fig.~\ref{fig:LAWNs}, the framework combines offline digital twin guidance with online world modeling in a hierarchical closed-loop architecture.

In LAWNs, constrained airspace, dense node deployment, and complex interference conditions motivate the use of digital twins for global situational awareness~\cite{Xie2023RadarIB,Wang2025ComputingPI}. By integrating virtual replicas of urban 3D models, building distributions, wireless propagation environments, and UAV operational states, digital twins support system-level optimization of UAV trajectory planning, task assignment, and resource scheduling. Xie et al.~\cite{Xie2023RadarIB} proposed a UAV application framework in which multiple tasks in the digital twin are coordinated by a task manager and interact with physical UAVs, enabling intelligent operation and management of real UAV networks. Wang et al.~\cite{Wang2025ComputingPI} developed a digital replica of aerial networks to jointly design power control, task partitioning, and computation resource allocation.

World models act as local embodied decision-makers at the UAV edge. They enable real-time prediction of channel dynamics and environmental disturbances to generate fine-grained control policies. Bar et al.~\cite{bar2025navworld} proposed a navigation world model based on the conditional diffusion transformer architecture, which predicts future visual observations conditioned on navigation actions, allowing UAVs to imagine flight trajectories in unfamiliar environments from a single input image and to perform online path planning under complex dynamic constraints. For more challenging dynamic obstacle avoidance tasks, Diehl et al.~\cite{Diehl2025DIODI} proposed the 4D occupancy flow world model, which estimates and decomposes scene occupancy flow from sparse laser radar observations, completes instance shapes, and predicts their temporal evolution. This predictive capability enables UAVs to anticipate the motion of low-altitude objects and make sub-second safe avoidance decisions while maintaining communication link stability, without relying on frequent synchronization with centralized digital twins. In addition, reconfigurable intelligent surfaces (RIS) introduce extra degrees of freedom for signal reflection in LAWNs. By learning implicit mappings among position, phase configuration, and channel quality, world models support end-to-end policy optimization without explicitly constructing analytical channel models~\cite{Chen2025HybridNA}.

In practical LAWNs deployments, digital twins provide offline or quasi-online trajectory planning and resource configuration from a global network perspective, offering macroscopic guidance that defines task boundaries and long-term objectives. World models operate onboard aerial nodes, generating real-time policies based on local perception and action feedback to dynamically adjust navigation maps and predefined trajectories~\cite{wu2025lowaltitudewirelessnetworkscomprehensive}. This collaborative architecture ensures both continuous network coverage and efficient resource utilization at the system level, while enabling edge nodes to autonomously adapt to transient environmental changes, forming a hierarchical control framework that combines global topology optimization with local link-level adaptation.

\section{ OPEN RESOURCES PROJECT}\label{open}
This section provides related open-source projects of digital twin, world model, and EGI across various fields.

\begin{table*}[htbp]
  \centering
  \footnotesize
  \caption{A summary of frameworks for digital twin, world model, edge general intelligence}
  \label{tab:framework_summary_wordstyle}
  \begin{tabular}{p{1.3cm}p{2.5cm}p{4.5cm}p{5.7cm}}
    \toprule  % 第一条线（顶线）
    \textbf{Field} & \textbf{Method} & \textbf{Characteristic} & \textbf{Related Resource Link} \\
    \midrule  % 第二条线（中线）
    \multirow{5}{*}{\makecell{Digital\\Twin}} & FaceChain~\cite{liu2023facechain} & Training-free and compatible & https://github.com/modelscope/facechain \\
    & PsyDT~\cite{xieetal2025psydt} & LLMs and psychological counseling & https://github.com/scutcyr/SoulChat2.0 \\
    & DTTD2~\cite{Huang2025dttd2} & Robust and object tracking & https://github.com/augcog/DTTD2 \\
    & DTaaS~\cite{talasila2024composable} & Management and services & https://github.com/INTO-CPS-Association/DTaaS \\
    & Ditto~\cite{jiang2022ditto} & PointNet++ and articulated object & https://github.com/UT-Austin-RPL/Ditto \\
    \hline
    \multirow{5}{*}{\makecell{World\\Model}} & LWM~\cite{liu2023LWM} & Context understanding and training & https://github.com/LargeWorldModel/LWM \\
    & DreamerV3~\cite{hafner2025nature} & RL and imagination-based planning & https://github.com/danijar/dreamerv3 \\
    & LingBot-World~\cite{Robbyant2026lingbotworld} & High-fidelity and long-horizon & https://github.com/Robbyant/lingbot-world \\
    & IRIS~\cite{iris2023worldmodels} & Data-efficient and Sequence-modeling & https://github.com/eloialonso/iris \\
    & GigaBrain-0~\cite{gigaai2025gigabrain0} & Policy robustness and spatial reasoning & https://github.com/open-gigaai/giga-brain-0 \\
    \hline
    \multirow{5}{*}{\makecell{Edge General\\Intelligence}} & LotteryFL~\cite{Li2021LotteryFL} & Personalized and Low-Comm FL & https://github.com/charleslipku/LotteryFL \\
    & Neurosurgeon~\cite{kang2017neurosurgeon} & Fine-grained and Layer-wise & https://github.com/Tjyy-1223/Neurosurgeon \\
    & FedCache~\cite{wu2024fedcache} & Device-Fit and Personalized & https://github.com/wuzhiyuan2000/FedCache \\
    & ORRIC~\cite{cai2024ORRIC} & Adaptive Inference and Retraining & https://github.com/caihuaiguang/ORRIC \\
    & pFedSD~\cite{Jin2023pFedSD} & Faster personalization and robustness & https://github.com/CGCL-codes/pFedSD \\
    \bottomrule  % 第三条线（底线）
  \end{tabular}
\end{table*}

%  Digital Twin
\subsection{Digital Twin Framework}
Digital twins have advanced in multiple dimensions, including generation, behavior modeling, robust tracking, platform support, and object reconstruction. The following work highlights representative frameworks and methods that illustrate progress in these areas.

\subsubsection*{Deep Learning Tools for Digital Twin Generation}
Deep learning has become a key tool for digital twin generation. In this context, \textit{FaceChain}~\cite{liu2023facechain} is a deep learning framework to generate human portraits that preserve identity. It uses decoupled training and face-related perceptual understanding models to extract ID features, combined with Classifier-Free Guidance and models such as DamoFD~\cite{liu2023damofd} and M2FP~\cite{Cheng2022M2FP}. The framework supports high-quality controllable portrait generation with custom style training and pose control, producing realistic outputs.

\subsubsection*{LLM-enhanced Digital Twin Frameworks}
Large language models provide new capabilities for building digital twins with personalized behaviors. \textit{PsyDT}~\cite{xieetal2025psydt} is an LLM-based framework to build digital twins of psychological counselors with personalized styles. It uses dynamic one-shot learning to capture counselor linguistic patterns and therapy techniques, synthesizing multi-turn dialogues to fine-tune the model for personalized counseling behavior.

\subsubsection*{Robust Digital Twin Tracking}
Robust pose estimation under sensor noise is essential for reliable digital twin tracking in mobile environments. \textit{DTTDNet}~\cite{Huang2025dttd2} addresses this problem by introducing a robust six-degrees-of-freedom (6DoF) pose estimation network for mobile environments. Built on a Transformer, it uses geometric feature filtering and Chamfer Distance loss to enhance robustness to depth noise. Experiments on DTTD-Mobile, a new digital-twin dataset from mobile devices, show that DTTDNet achieves 60.74 on the ADD metric, outperforming existing methods by at least 4.32, and remaining stable across noise levels.

\subsubsection*{Composable and Reusable Digital Twin Platform}
Platform-based solutions reduce complexity and enhance reusability in digital twin systems. \textit{DTaaS}~\cite{talasila2024composable} offers a composable platform that centrally manages assets—models, data, functions, and tools—and supports building digital twins as services, integrating storage, computing, communication, monitoring, and task execution. Case studies show its effectiveness for development and service-oriented deployment.

\subsubsection*{Digital Twins for Articulated Object Reconstruction and Modeling}
Recent studies have explored digital twins to model the structure and motion of articulated objects. \textit{Ditto}~\cite{jiang2022ditto} constructs digital twins of real-world articulated objects through interactive perception. Using visual observations before and after interaction, it reconstructs part-level geometry and estimates articulation models with implicit neural representations. The method is category-agnostic and supports real-world reconstruction and physical simulation.

These works demonstrate advances in digital twins across generation, personalized behavior, robust tracking, platform-based management, and modeling of complex objects. They improve the accuracy, controllability, and reusability of digital twins, supporting the application of EGI in perception, reasoning, and autonomous decision-making.

%   World Model
\subsection{World Model Framework}
World models provide a framework for representing and predicting complex environments. They combine multimodal perception, decision-making, planning, and simulation capabilities. These models help agents act and learn efficiently across long tasks, high-dimensional spaces, and multiple domains, supporting applications in AI and robotics.

\subsubsection*{LLM-enhanced World Model Frameworks}
Multimodal models with extended context have shown significant progress in recent years. Building on these advances, \textit{LWM}~\cite{liu2023LWM} enables cross-modal understanding and generation of text, images, and videos while handling long-context inputs. It excels in long-text retrieval, long-video understanding, and text-to-image/video tasks, while retaining short-context capabilities and providing an open-source training pipeline.

\subsubsection*{World Model-Driven RL for Multi-Domain Tasks}
General-purpose reinforcement learning (RL) has advanced rapidly in recent years. \textit{DreamerV3}~\cite{hafner2025nature} is a world model-based algorithm with robustness techniques such as normalization, Kullback-Leibler (KL) divergence balancing, and symlog/symexp transformations. It adapts to 150+ tasks across eight domains without tuning, learns efficiently via unsupervised reconstruction and actor–critic methods, outperforms proximal policy optimization (PPO) and MuZero~\cite{schrittwieser2020mastering} algorithm on benchmarks, and offers flexible model size and replay ratio for practical cross-domain use.

\subsubsection*{Cross-Domain Applications of World Models}
Interactive simulation platforms have advanced rapidly with video generation and world modeling. \textit{LingBot-World}~\cite{Robbyant2026lingbotworld} provides an open-source platform for multi-domain simulations. It integrates a hierarchical semantic engine, multi-stage training, and mixture-of-experts (MoE) architecture to deliver high-fidelity, long-term, low-latency environments, supporting prompt-driven events, agent training, and 3D reconstruction.

\subsubsection*{Visual and Long-Horizon World Models}
Recent RL research has focused on world model-driven approaches for sample-efficient learning. \textit{IRIS}~\cite{iris2023worldmodels} builds a world model with a discrete autoencoder and autoregressive Transformer for learning in imagination. Trained on simulated trajectories with real data, it performs pixel-level prediction, reward and termination estimation, adapts to complex visual environments, and outperforms humans in Atari 100k games.

\subsubsection*{Efficient Robot Learning with World Models}
Efficient robot learning depends on scalable data and robust task generalization. \textit{GigaBrain-0}~\cite{gigaai2025gigabrain0} uses data generated by world models to build a vision-language-action (VLA) foundation model, reducing the reliance on real robot data. With RGB-depth-map (RGBD) modeling and embodied chain-of-thought supervision, it reasons about geometry, object states, and long-horizon dependencies, achieving robust performance in dexterous and mobile tasks.

These studies highlight the role of world models in supporting EGI. By enabling reasoning, planning, simulation, and robot learning, they improve efficiency and generalization, supporting robust perception, decision-making, and autonomous action in complex, multi-domain environments.

%   EGI Framework
\subsection{Edge Artificial Intelligence Framework}
EGI faces the challenge of efficient, personalized learning and inference across distributed heterogeneous devices. Integrating world models, personalized federated learning, resource-aware scheduling, and knowledge distillation improves efficiency, generalization, and supports diverse intelligent applications.

\subsubsection*{EGI with Communication-Efficient Federated Learning}
Reducing communication and enabling personalization are critical for federated learning on edge devices. \textit{LotteryFL}~\cite{Li2021LotteryFL} uses the Lottery Ticket hypothesis to train and transmit client-specific subnetworks, lowering communication overhead while supporting personalized models. Experiments on non-identically independently distributed (non-IID) datasets show improved accuracy and efficiency, with real-time deployment demonstrated on edge devices.

\subsubsection*{Resource-Aware EGI Frameworks}
EGI applications require low latency and high energy efficiency. To address these challenges, \textit{Neurosurgeon}~\cite{kang2017neurosurgeon} uses layer-level neural network partitioning to coordinate edge and cloud computing resources. The framework predicts performance and dynamically adapts to hardware, architecture, and network conditions, achieving improved efficiency and throughput.

\subsubsection*{Communication-Efficient FL for EGI}
Efficient and personalized learning is critical for edge devices. \textit{FedCache}~\cite{wu2024fedcache} uses a knowledge cache on the server to provide relevant information to client models and applies ensemble distillation. The framework supports heterogeneous devices and asynchronous interactions, achieving significant communication efficiency gains while maintaining performance comparable to state-of-the-art methods.

\subsubsection*{Balancing Model Drift and Inference in EGI}
Managing computation and maintaining accuracy are key to practical edge intelligence. In response, \textit{ORRIC}~\cite{cai2024ORRIC} models resource competition between retraining and inference and dynamically adapts resource allocation. It improves long-term inference accuracy while balancing drift-related losses and optimizing computational resources and latency.

\subsubsection*{Self-Distilled FL for Edge Devices}
Edge computing applications require efficient and personalized model training. \textit{pFedSD}~\cite{Jin2023pFedSD} is a personalized federated learning framework for edge computing. It uses self-knowledge distillation to retain client historical models and guide local training, improving personalization and convergence while supporting non-IID data, heterogeneous models, and preserving privacy with low system overhead.

These studies highlight the importance of communication-efficient, resource-aware, and personalized strategies in edge environments. By reducing overhead, managing model drift, optimizing inference, and supporting diverse devices, they provide a solid basis for efficient and reliable EGI systems.

\section{Future Directions}\label{future}
Advancing world models for EGI requires realism, adaptability, and efficiency in dynamic and constrained environments. Edge systems need models that combine physical knowledge with data-driven learning, support continual adaptation, and operate at different spatial and temporal scales. The following directions outline key areas for developing robust, explainable, and cooperative world models.  

\subsection{Hybrid Physics-Driven and Data-Driven World Models}
Future work should investigate hybrid architectures that combine explicit physical knowledge (e.g., radio propagation models, mobility laws, power constraints) with learned latent dynamics. Purely data-driven world models often struggle with extrapolation under sparse, non-stationary, or shifted data, and their generalization and scalability in complex real-world robotic scenarios remain in question~\cite{zhang2025whale}. In contrast, purely physics-based digital twins can be brittle and computationally expensive. Promising directions include physics-informed latent spaces, generative models regularized by conservation laws~\cite{utkarsh2025physicsconstrained}, and differentiable simulators coupled with neural world models. For EGI, such hybrid designs can improve robustness and interpretability while remaining lightweight enough for deployment at the edge.

\subsection{Federated World Modeling at the Edge}
EGI requires world models that evolve as environments, traffic patterns, and hardware conditions change. Future research should address continual and lifelong learning of world models under strict resource and privacy constraints, so that intelligent systems can retain existing knowledge while continuously acquiring and integrating new information~\cite{Zheng2026lifelongLearning}. This includes mechanisms for online adaptation without catastrophic forgetting, efficient model versioning across heterogeneous edge nodes, and federated learning protocols tailored to world-model training (e.g., regularizing latent dynamics to improve agent behavior~\cite{Saanum2024latentdynamics}). Handling non-IID data, enabling communication-efficient aggregation, and supporting privacy-preserving updates will be central challenges.

\subsection{Multi-Agent and Multi-Scale World Models}
Edge scenarios such as 6G, air-to-ground networks, and low-altitude operations inherently involve many interacting agents (devices, base stations, UAVs, vehicles) evolving across multiple temporal and spatial scales. Future research should explore world models that capture multi-agent interactions (e.g., via graph-structured agent-level interaction modules~\cite{Jeong2024multiagent}) and multi-scale processes (fast wireless-channel fluctuations vs. slower mobility and traffic patterns). Agent-centric models with interactive perception capabilities can support decentralized collaboration among distributed edge nodes, enable predicting emergent behaviors, and facilitate cooperative planning. Decentralized agents may form collusion through covert communication~\cite{motwani2024secret}, highlighting the need to move beyond single-agent, local-view world models and to develop mechanisms for effective decentralized planning.

\subsection{Explainable and Trustworthy World Models}
As EGI systems are deployed in high-stakes environments, it is crucial that their decisions are understandable, reliable, and trustworthy. Current world models, often based on deep neural networks, tend to act as black boxes, making failures difficult to interpret and diagnose~\cite{Aghababaeyan2023deepneuralnetworks}. A key research direction is the development of explainable artificial intelligence (XAI) techniques tailored to world models. Potential approaches include methods to visualize the model's imagined futures, to identify which environmental features are most influential for its predictions, and to quantify uncertainty in its forecasts. Building trust also requires mechanisms to detect out-of-distribution conditions and to decide when the world model should not be used for planning, as such detection is a key component of trusted machine learning systems~\cite{Zheng2025Trust} and enables graceful degradation, conservative fallback strategies, or safe human or rule-based intervention.

\section{Conclusion}
This survey has outlined a unified perspective on the transition from digital twins to world models for EGI. Digital twins remain indispensable for high-fidelity engineering analysis, lifecycle management, and system-level optimization. However, their reliance on explicit modeling, centralized computation, and continuous synchronization limits their suitability for autonomous and real-time operations at the edge. World models address these limitations by learning compact, action-conditioned representations of the environment and by enabling imagination-based planning and self-supervised adaptation. This survey has reviewed core architectures and algorithms for world models, their coupling with communication, sensing, and control, and their emerging role in future wireless networks and cyber–physical systems. We have also highlighted open issues, including hybrid physics–data integration, federated and continual world modeling under non-IID edge data, multi-agent and multi-scale modeling, as well as safety, explainability, and standardization. These challenges define a rich research agenda for the coming years. Thus, the synergistic use of digital twins and world models is expected to provide a key technological pillar for robust, efficient, and intelligent edge systems in 6G and beyond.

\bibliography{Ref}

\end{document}